\pdfoutput=1

\documentclass[11pt]{article}

\usepackage{naacl2021}

\usepackage{times}
\usepackage{latexsym}

\usepackage[T1]{fontenc}

\usepackage[utf8]{inputenc}

\usepackage{microtype}
\usepackage{algorithm} 
\usepackage{algpseudocode}
\usepackage{amsmath}
\usepackage{amssymb}
\usepackage{graphicx}
\usepackage{booktabs,tabularx,enumitem,ragged2e}
\usepackage{comment}
\usepackage{framed}
\usepackage[normalem]{ulem}
\usepackage{xcolor}
\title{\textsc{multiPRover}: Generating Multiple Proofs for Improved Interpretability in Rule Reasoning}

\author{Swarnadeep Saha \quad Prateek Yadav \quad Mohit Bansal
\\ 
  UNC Chapel Hill\\ 
  \texttt{\{swarna, prateek, mbansal\}@cs.unc.edu}
}

\begin{document}

\newcommand{\model}{\textsc{multiPRover}}
\newcommand{\shortmodelmult}{ML-\textsc{multiPRover}}
\newcommand{\shortmodelseq}{IT-\textsc{multiPRover}}
\newcommand{\modelmult}{\textit{Multilabel}-\textsc{multiPRover}}
\newcommand{\modelseq}{\textit{Iterative}-\textsc{multiPRover}}

\newcommand{\mb}[1]{\mathbb{#1}}
\newcommand{\mc}[1]{\mathcal{#1}}
\newcommand{\sqb}[1]{\left[#1\right]}
\newcommand{\cirb}[1]{\left(#1\right)}
\newcommand{\curb}[1]{\left \{#1\right\}}

\maketitle
\begin{abstract}
We focus on a type of linguistic formal reasoning where the goal is to reason over explicit knowledge in the form of natural language facts and rules \cite{clark2020transformers}. A recent work, named \textsc{PRover} \cite{saha2020prover}, performs such reasoning by answering a question and also generating a proof graph that explains the answer. However, compositional reasoning is not always unique and there may be multiple ways of reaching the correct answer. Thus, in our work, we address a new and challenging problem of generating multiple proof graphs for reasoning over natural language rule-bases. Each proof provides a different rationale for the answer, thereby improving the interpretability of such reasoning systems. In order to jointly learn from all proof graphs and exploit the correlations between multiple proofs for a question, we pose this task as a set generation problem over structured output spaces where each proof is represented as a directed graph. We propose two variants of a proof-set generation model, \model{}. Our first model, \modelmult, generates a set of proofs via multi-label classification and implicit conditioning between the proofs; while the second model, \modelseq{}, generates proofs iteratively by explicitly conditioning on the previously generated proofs. Experiments on multiple synthetic, zero-shot, and human-paraphrased datasets reveal that both \model{} models significantly outperform \textsc{PRover} on datasets containing multiple gold proofs. \modelseq{} obtains state-of-the-art proof F1 in zero-shot scenarios where all examples have single correct proofs. It also generalizes better to questions requiring higher depths of reasoning where multiple proofs are more frequent.
\end{abstract}

\begin{figure*}
\centering
    \includegraphics[width=0.99\textwidth]{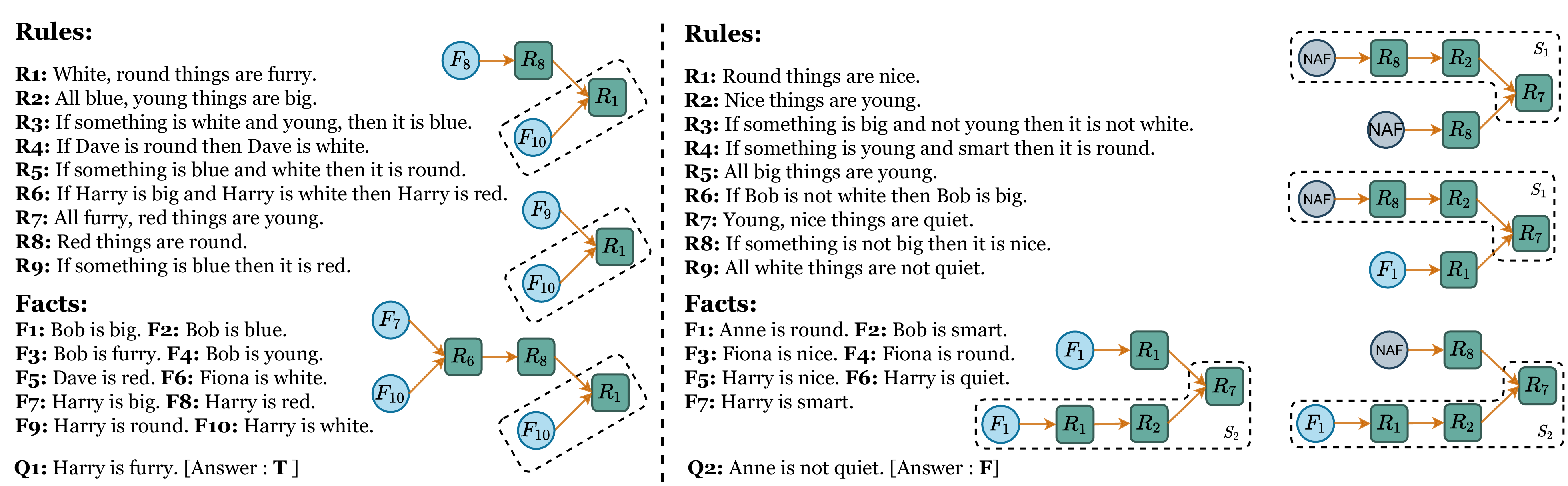}
    \vspace{-6pt}
\caption{Two rule-bases with rules, facts, questions, answers and all possible proofs. The first question has three correct proofs while the second question has four correct proofs. \model{} answers both questions correctly and also generates all the corresponding proofs accurately for each question.
\vspace{-10pt}}

\label{fig:example_dataset_proof}
\end{figure*}

\section{Introduction}

Formal reasoning over explicit multi-sentence knowledge \cite{newell1956logic} has often proved to be challenging \cite{musen1988brittleness}, owing to the difficulty in creating logical forms from such sentences, thereby restricting the application of semantic parsers ~\cite{zettlemoyer2012learning, berant2013semantic, berant2014semantic}. Thus, in a recent work, \citet{clark2020transformers} bypass the creation of intermediate logical forms and show that transformers \cite{vaswani2017attention} can act as ``soft theorem provers" by answering questions over natural language (English) rule-bases, consisting of facts and rules. In order to reliably interpret these predicted answers, \citet{saha2020prover} propose \textsc{PRover}, a transformer-based model that generates the corresponding proof graph, thus emulating formal reasoning closely. Consider the two example rule-bases with two questions and corresponding proofs in Figure \ref{fig:example_dataset_proof}, where a proof is a directed graph consisting of the relevant facts and rules from the corresponding rule-base. 

\textsc{PRover} shows good single-proof generation accuracy but is designed and trained in a way to generate only a single proof for each question.
This is not ideal because formal proofs are not always unique and there may be multiple correct ways of arriving at the answer. For example, $Q_1$ and $Q_2$ in Figure \ref{fig:example_dataset_proof} have three and four correct proofs respectively. Hence, in order to enhance the human-interpretability of linguistic formal reasoning systems, it is desirable to develop methods that can generate multiple proofs, each providing a different rationale for the predicted answer. Such interpretable methods, while possessing the flexibility of operating over natural language, can also aid in verifying claims when constructing proofs from scratch is tedious or infeasible. 

We find that \textsc{PRover} \cite{saha2020prover}, when trained on all proofs as independent training examples (Eq. \ref{eqn:dataset}) and extended to generate top-$p$ proofs during inference (Eq. \ref{eqn:node_inferenc}), fails drastically, achieving a low proof precision of 34\%. The subsequent proofs are often incorrect because it is not trained jointly with all proofs and hence, is unable to exploit the inter-proof correlations and also does not learn the correct number of proofs for a question. Thus, we propose \model{}, a transformer-based model that can generate a set of proof graphs with appropriate cardinality for a given question. Since multiple proofs can be generated in any arbitrary order, we pose this task as a set generation problem over graphs and train \model{} jointly with a permutation-invariant Hungarian Loss \cite{Zhang2019deepsets, zhang2019fspool} over all proofs.

A proof graph is generated through a node module which selects the relevant facts and rules as part of the proof and an edge module which determines the edges between the chosen nodes. Similar to PRover, we first enforce multiple structural constraints during training and inference to ensure that a generated proof is valid. Next, in order to generate a set of proofs jointly, we propose our first model, \modelmult{}, a multi-label classification framework which performs \textit{implicit conditioning} among the proofs and predicts $p$ binary labels for each node and edge, denoting its presence or absence in each of the $p$ proofs that we want to generate. It is efficient in terms of the number of parameters and training time and also achieves a better proof F1 than \textsc{PRover}. However, the lack of explicit conditioning between the proofs is not ideal because a question with multiple proofs often has certain common sub-graphs across the proofs. E.g., all the 3 proofs for $Q_1$ in Figure \ref{fig:example_dataset_proof} have the sub-graph $\{F_{10} \rightarrow R_1\}$ common. Thus, in order to exploit these correlations which \modelmult{} cannot capture explicitly, we further propose an improved variant of \model{}, named \modelseq{}, which generates an appropriate number of proofs by stacking multiple node and edge encoders, each of which generates one proof at each time step by conditioning on the previously generated proofs. This enables the model to better learn the correlations between multiple proofs for a given question. To capture the set-based nature of the task, we train \model{} using a permutation-invariant Hungarian Loss (Sec.~\ref{sec:hungarian}), which solves an assignment problem between a set of predicted and gold proofs. 

Empirical evaluation on synthetic and human paraphrased QA rule-bases \cite{clark2020transformers} show that both of our \model{} models achieve a significantly higher proof F1 compared to \textsc{PRover} while retaining the QA accuracy. Further, on a challenging hand-authored zero-shot dataset, where all examples have single gold proofs, \modelseq{} achieves state-of-the-art proof F1. It also generalizes better to questions requiring higher depths of reasoning with more multiple proofs.
Overall, our contributions are:
\begin{itemize}[nosep, wide=0pt, leftmargin=*, after=\strut]
    \item We address a new and challenging problem of generating a set of multiple logical proof graphs for reasoning over natural language rule-bases by proposing two set-based joint models, \modelmult{} and \modelseq{}.\footnote{Our code and models are publicly available at \url{https://github.com/swarnaHub/multiPRover}.}
    \item \modelseq{}'s joint training and explicit conditioning helps it to better learn the relative importance of rules and facts for a particular question and uncover common subgraphs across multiple proofs. Thus, compared to \modelmult{} and \textsc{PRover}, it is able to transfer well in zero-shot settings because it learns to assign a soft prior over the rule-base.
    \item \modelseq{}'s conditional generation also enables it to generalize better to questions requiring higher depths of reasoning where the presence of multiple proofs is frequent.
\end{itemize}{}

\section{Related Work}

The task of rule reasoning \cite{clark2020transformers} is related to other recently proposed tasks on QA \cite{weston2015towards, yang2018hotpotqa, lin-etal-2019-reasoning, tafjord-etal-2019-quartz, richardson2020probing} and NLI \cite{maccartney2014natural}. However, most of these tasks require implicit reasoning rules as opposed to explicit ones and the focus is either on broad language understanding or on single rule application. Below we discuss \model{}'s relation to multiple areas of NLP and ML.

\paragraph{Structured Explanations:} There is useful previous work on developing interpretable and explainable models \cite{doshi2017towards, rudin2019stop,hase_evaluating_2020, jacovi2020towards} for NLP. Explanations in NLP take three major forms -- (1) extractive rationales or highlights \cite{zaidan2007using, lei2016rationalizing, yu2019rethinking, deyoung2020eraser} where a subset of the input text explain a prediction, (2) free-form or natural language explanations \cite{camburu2018snli, rajani2019explain,zhang2020WinoWhy, kumar2020nile} that are not constrained to the input, and (3) structured explanations that range from semi-structured text \cite{ye2020teaching} to chain of facts \cite{khot2020qasc, jhamtani2020learning,gontier2020measuring} to explanation graphs (based on edges between chains of facts)  \cite{jansen2018worldtree, jansen2019textgraphs, xie2020worldtree}.

\paragraph{Generating Multiple Outputs:} Generating a set of proofs can be viewed as a task of generating multiple structured outputs \cite{prasad2014}. Multiple prior studies focus on generating diverse unstructured texts \cite{gimpel2013, dai2017towards, xu2018d, raffel2019t5}. which broadly span two categories -- (1) using improved decoding techniques like beam search with inter-sibling ranking penalty \cite{li2016simple}, iterative beam search \cite{kulikov2018importance}, diverse beam search \cite{vijayakumar2016diversebeam}, and sentence codes \cite{shu2019generating}, (2) varying the hidden representations or using multiple decoders \cite{dai2017towards, jain2017creativity, shen2019mixture}. Our baseline, ~\textsc{PRover}-top-$p$, which extends \textsc{PRover} to generate top-$p$ proofs during inference falls in the first category while \model{} falls in the second category, where the multiple node and edge encoders vary the node and edge representations for generating multiple proofs.

\paragraph{Machine Learning over Sets:} Set-based ML models \cite{Zaheer2017deepsets, lee2018settoset, Zhang2019deepsets,kosiorek2020conditional} have a wide range of applications including generating multiple image captions \cite{vinyals2014caption}, generating diverse translations \cite{cho2014mt,bahdanau2014neural}, enumerating rules in a logical inference system \cite{gao2019sequential}. Set problems are challenging because the number of valid solutions for a set of size $n$ are $n!$, which increases faster than exponential in $n$ and ignoring the set structure produces sub-optimal solutions \cite{Zhang2019deepsets}. Thus, we use a set-based Hungarian Loss \cite{Zhang2019deepsets, zhang2019fspool} for capturing the permutation-invariant nature of generating a set of proofs.

\section{Method} 

\subsection{Task Description and Notations}
\label{sec:task}
The input to our task is a tuple of the form $(\mc{C}, \mc{Q})$, where $\mc{C}$ is a rule-base context and $\mc{Q}$ is the question. We want to predict a binary answer $\mc{A} \in \{True,False\}$ for the question and generate a set of proof graphs $P = \{\mc{P}_1, \hdots, \mc{P}_p\}$, each of which provides a diverse rationale for the answer (see Figure \ref{fig:example_dataset_proof}). 
The context $\mc{C}$ consists of a set of facts and rules, denoted by $\mc{F}$ and $\mc{R}$ respectively. Facts $\mc{F} = \{F_1, \hdots F_f\}$ are unambiguous statements, while rules $\mc{R} = \{R_1,  \hdots R_r\}$ are logical statements, which can be used in conjunction with the facts to arrive at a logical conclusion. Each proof $\mc{P}_i = (\mc{V}_i, \mc{E}_i)$ is a directed graph, with a set of nodes $\mc{V}_i \subseteq \mc{N}$ and a set of edges $\mc{E}_i \subseteq \mc{V}_i \times \mc{V}_i$, where $\mc{N} = \mc{F} \cup \mc{R} \cup \{\textbf{NAF}\}$ and $k = |\mc{N}|$. If a statement (E.g. ``Anne is big'') cannot be deduced from the context, then Negation as Failure ($\textbf{NAF}$) contains the negation of that statement (E.g. ``Anne is not big''), which is considered true in a closed-world assumption. See appendix for more details of the syntax of proof graphs.

\subsection{Baseline \textsc{PRover} Model}
\label{sec:prover}

\textsc{PRover} \cite{saha2020prover} builds on top of RoBERTa \cite{liu2019roberta} and consists of a question answering (QA) module, a node module, and an edge module where the node and edge modules are used to predict a single proof graph. The input to RoBERTa is the concatenation of the facts, rules, and the question. The QA module takes in the representation of the $[CLS]$ token and predicts a binary label for the question. The node module computes the node embeddings $\textbf{N} \in \mb{R}^{k \times d}$ consisting of the representations of each fact, rule and \textbf{NAF} where $d$ is the embedding dimension. The $i^{th}$ row $n_i$ of $\textbf{N}$ denotes the embedding of node $i$. A node classifier takes in these embeddings to output the node probabilities $np_i \in \mb{R}^k$ for each fact, rule, and \textbf{NAF} being present in the proof. The edge module computes the edge embeddings $\textbf{E} \in \mb{R}^{k^2 \times 3d}$ for every edge $(i, j)$ through the function $\phi(i,j) = [n_i; n_j; (n_i-n_j)]$ where $;$ is the concatenation operation and outputs probabilities $ep_{i,j} \in \mb{R}^{k^2}$ of each edge being present in the proof. \textsc{PRover} is trained using the joint cross-entropy loss over the QA, node, and edge modules. The authors pose inference as an Integer Linear Program (ILP). Given a set of nodes and the edge probabilities from the trained model, the following global score over the edge probabilities is maximized, subject to multiple structural constraints $\mc{S}$ that ensure the validity of a proof graph (like checking for graph connectivity).
\begin{equation}
\small
    \underset{e_{i,j}\in \{0,1\},\\ s \in \mc{S}}{\operatorname{\arg\max}} \sum_{i,j,i \neq j} ep_{i,j} * e_{i,j} + (1-ep_{i,j}) * (1-e_{i,j})    
    \label{eqn:edge_opt}
\end{equation}

\begin{figure}
    \centering
    \includegraphics[width=\columnwidth]{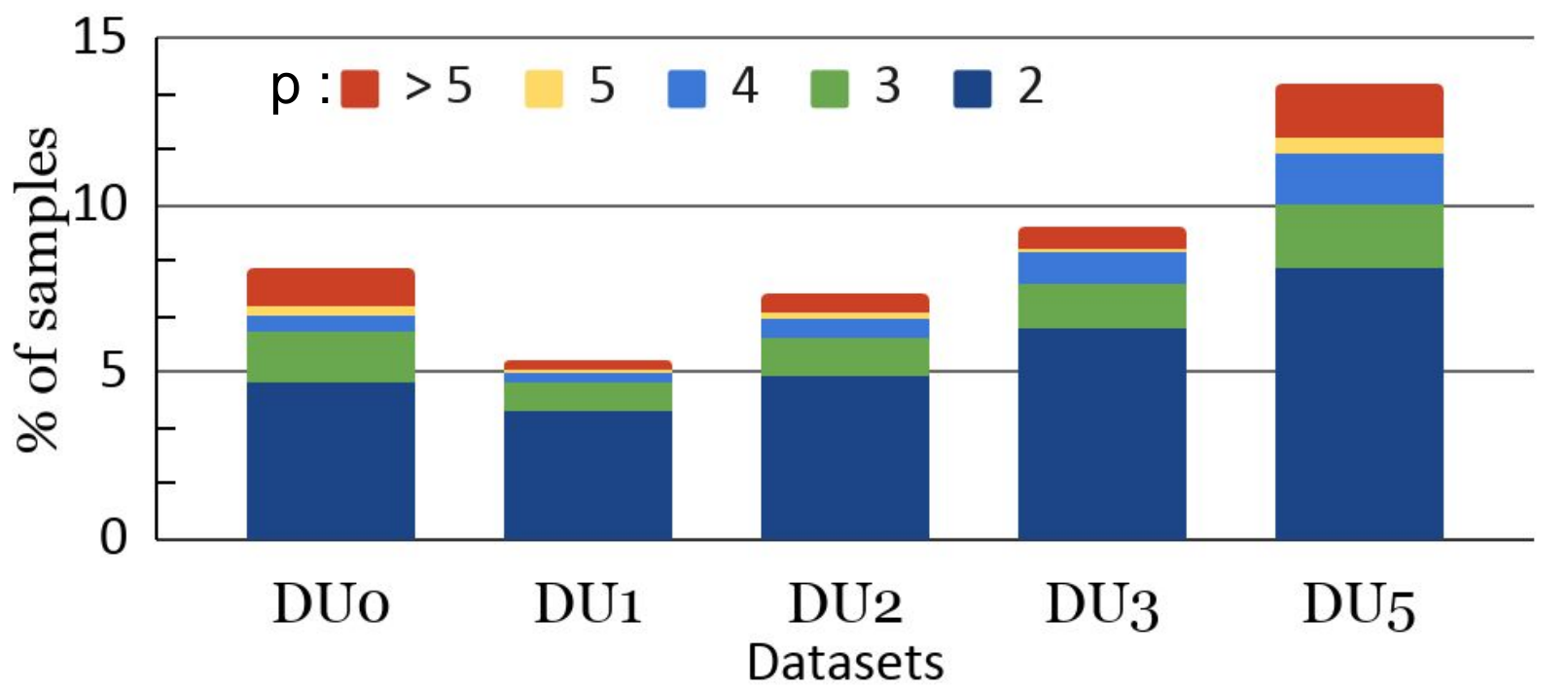}
    \vspace{-15pt}
    \caption{Plot showing the percentage of samples with $p>1$ proofs for different training datasets, DU0-DU5.}
    \vspace{-8pt}
    \label{fig:percent_proof}
\end{figure}

\noindent \textbf{Extending \textsc{PRover} to Generate Proof-Sets:} Since \citet{saha2020prover} focus on generating one proof per question, they also train their model with one gold proof per question. For multiple proof generation, an obvious extension is to treat each proof for a question as a separate training example. Formally, for each sample $l$, given a context $\mc{C}^l$, a question $\mc{Q}^l$, an answer $\mc{A}^l$  and a set of gold proofs $\mc{P}^{l}_{i}$, where $i \in \{1,\hdots, p_l\}$, the extended training dataset can be defined as:
\begin{equation}
    \footnotesize
    \mc{D} = \bigcup_{l=1}^{L} \curb{\cirb{\mc{Q}^l, \mc{C}^l, \mc{A}^l, \mc{P}^l_i}_{i=1}^{p_l}}_{l}
    \label{eqn:dataset}
\end{equation}

Once \textsc{PRover} is trained with this dataset, during inference, we generate top-$p$ proofs by first selecting the top-$p$ node sets according to Eqn.~\ref{eqn:node_inferenc} and then choosing the corresponding edge sets using the optimization function in Eqn.~\ref{eqn:edge_opt}. 
\begin{equation}
    \underset{\textbf{v} \in \{0,1\}^k} {\operatorname{\arg\max}}
    \sum_{i=1}^k np_{i} * \textbf{v}_{i} + (1-np_{i}) * (1-\textbf{v}_{i})
    \label{eqn:node_inferenc}
\end{equation}

\begin{figure}
    \centering
    \includegraphics[width=\columnwidth]{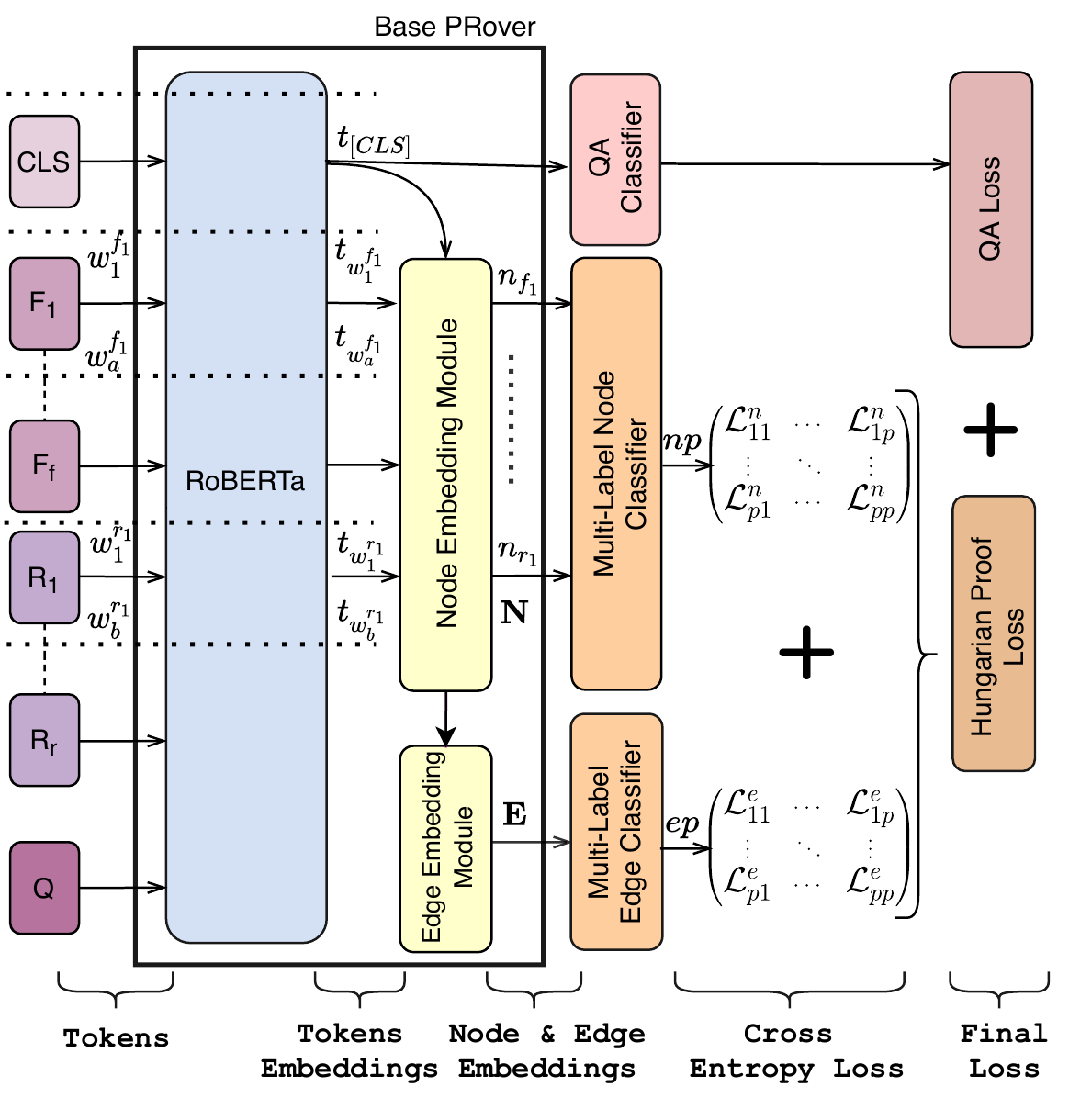}
    \vspace{-20pt}
    \caption{\modelmult{}.}
    \vspace{-10pt}
    \label{fig:ml-mprover}
\end{figure}

The top-$p$ solutions of Eqn.~\ref{eqn:node_inferenc} are $\textbf{v}^1, \hdots, \textbf{v}^p$ which indicate a node's presence or absence in the proofs. Although simple, this approach has two major issues. First, the lack of coupling between the proofs can potentially confuse the model as there are multiple possible proofs for the same (question, context) pair. Second, inference is inflexible and always generates a fixed number of proofs for every example, thus leading to the generation of many incorrect proofs (Section \ref{exp:main}). As shown in Figure \ref{fig:example_dataset_proof}, certain questions can have multiple possible proofs. Figure \ref{fig:percent_proof} demonstrates this phenomenon statistically -- the datasets we experiment with \cite{clark2020transformers} contain up to 13\% of the samples with $>$ 1 correct proof. Thus, in the light of \textsc{PRover}'s limitations, we propose two novel architectures of a proof-set generation model, \model{}.

\begin{figure}
    \centering
    \includegraphics[width=\columnwidth]{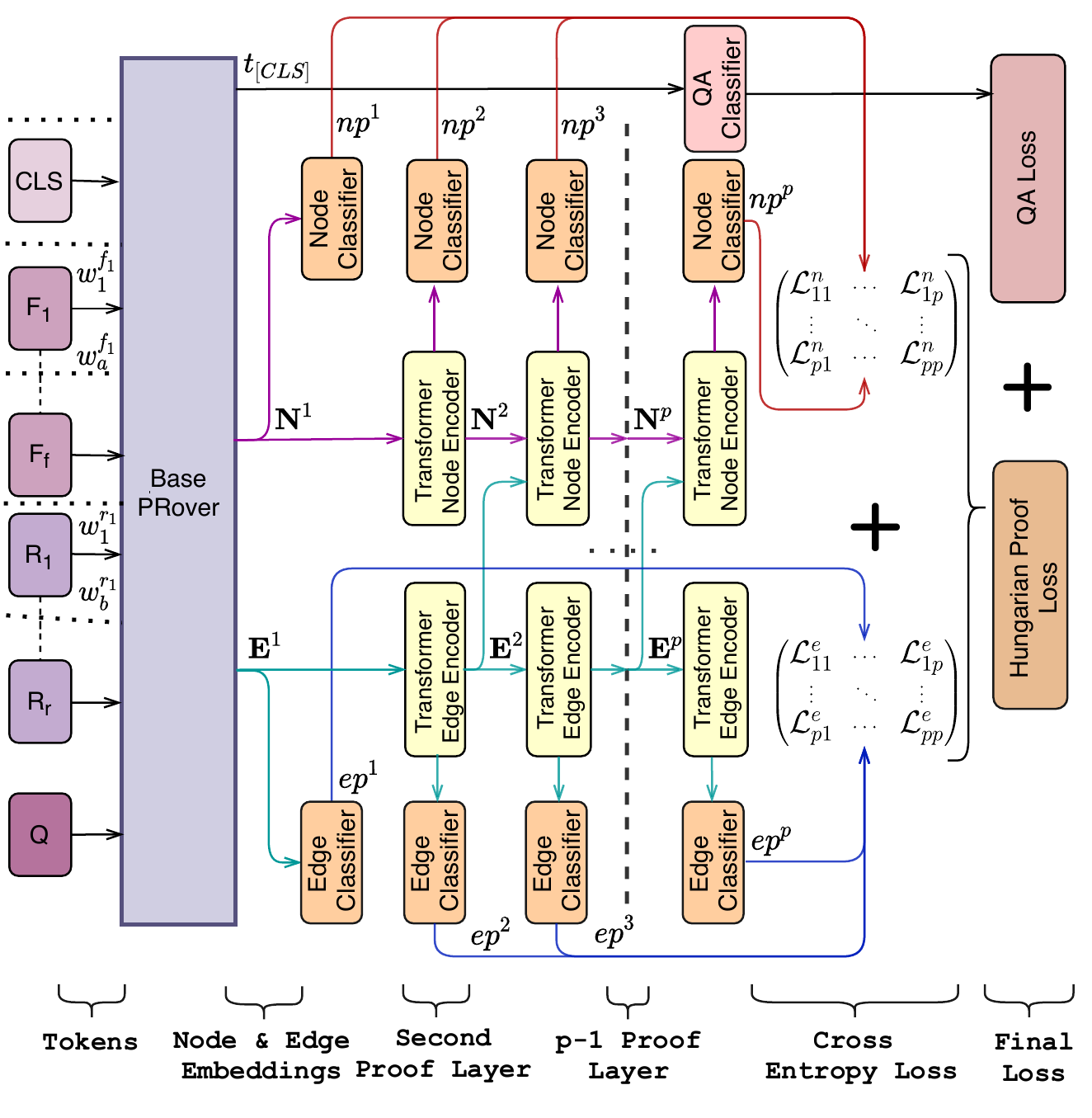}
    \caption{\modelseq{}.\vspace{-5pt}}
    \vspace{-5pt}
    \label{fig:it-mprover}
\end{figure}

\subsection{\modelmult}
\label{mod:mult}
As described in the previous section, a desired property for generating a set of proofs is to have the proofs conditioned on each other as opposed to treating them independently. Thus, we propose \modelmult{} (see Figure \ref{fig:ml-mprover}), which poses the problem of generating a set of proofs as a multi-label classification task over all the nodes and edges corresponding to the set of $p$ proofs. Each training example is a tuple $\cirb{\mc{Q}^l, \mc{C}^l, \mc{A}^l, \{\mc{P}^l_i\}_{i=1}^{p_l}}$, consisting of a set of gold proofs $\{\mc{P}^l_i\}_{i=1}^{p_l}$ per example. It consists of a QA module, a node module, and an edge module. Following \textsc{PRover} (Section \ref{sec:prover}), we obtain the node representations $\textbf{N} \in \mb{R}^{k \times d}$ by mean-pooling over the constituent RoBERTa representations. These are then passed through a \textit{multi-label node classifier}, which consists of two linear layers and produces the probabilities $np_{i} \in \mb{R}^{p}$ of a node being present in the $p$ proofs. The node embeddings $n_i$ and $n_j$ for a pair of nodes are transformed by the function $\phi(i,j)$, described in Section \ref{sec:prover}, to output the edge embeddings $\textbf{E} \in \mb{R}^{k^2 \times 3d} $. We also have a \textit{multi-label edge classifier}, which takes in the edge embeddings to generate the probabilities $ep_{i,j} \in \mb{R}^{p}$ of an edge $(i,j)$ being present in the $p$ proofs. Lastly, a \textit{question answering} module predicts a binary answer for the question. Following \textsc{PRover}, during training, we mask certain impossible edges like fact to fact, rule to fact, and non-nodes. Given the outputs from the three modules, we train our model jointly over all proofs using a set-based Hungarian Loss. 

This model is advantageous because there is \textit{implicit conditioning} between the proofs as all the proofs are generated in parallel from the same node embeddings and edge embeddings. Thus, it has no additional time or memory overhead while also generating proof-sets better than \textsc{PRover} (Section \ref{exp:main}). However, it suffers from two major drawbacks. First, since the proofs are generated in parallel, the model is trained by padding empty proof graphs. Hence for higher values of $p$, the model has to learn more empty proofs, which makes the learning problem harder. Second, the proofs are not explicitly conditioned on each other. This motivates us to propose \modelseq{}.

\begin{figure}[t]
    \centering
    \includegraphics[width=\columnwidth]{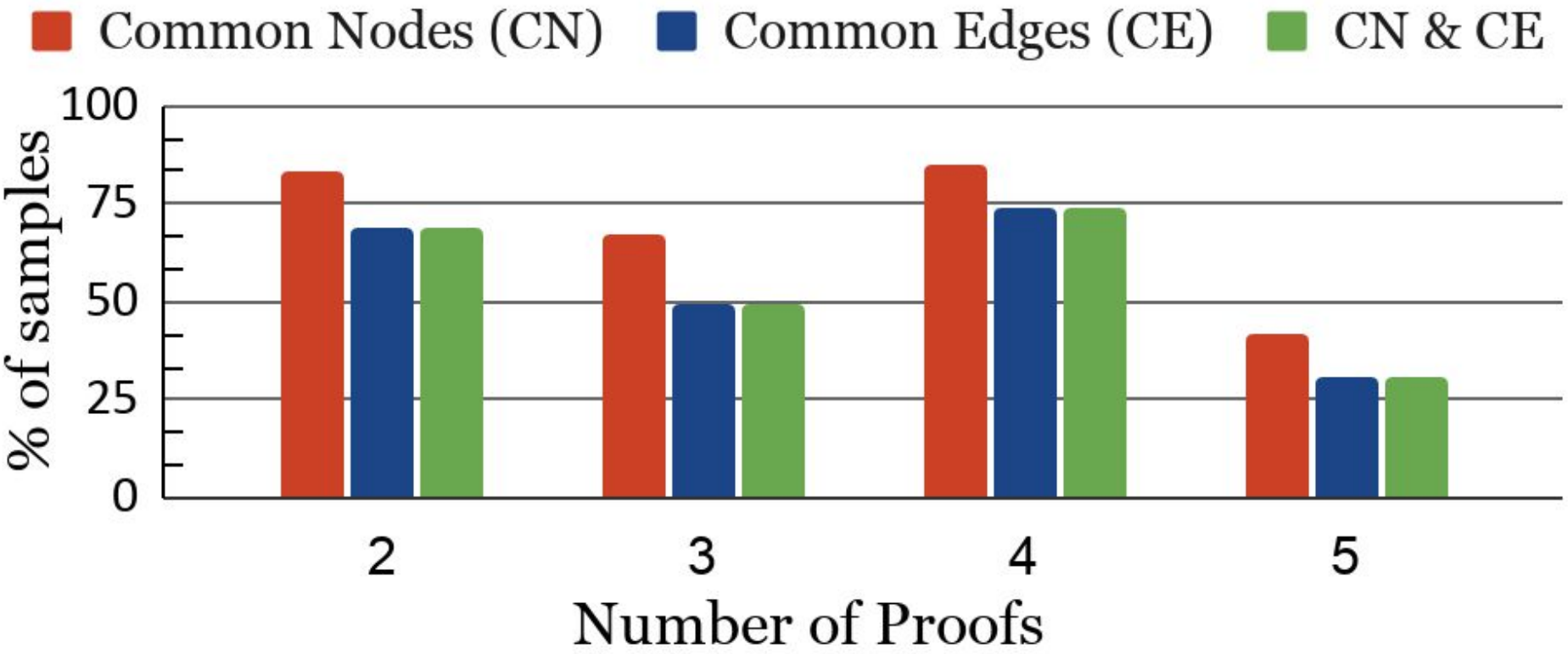}
    \vspace{-15pt}
    \caption{Plot showing the percentage of samples in DU5 with at least one common node, common edge or both between the proofs for varying number of proofs. \vspace{-2pt}}
    \vspace{-10pt}
    \label{fig:cnce}
\end{figure}

\subsection{\modelseq}
\label{mod:seq}
As a motivating example for why explicit conditioning among proofs is necessary, consider the proofs for $Q_1$ in Figure \ref{fig:example_dataset_proof} where the sub-graph $\{F_{10} \rightarrow R_1\}$ is common across all the proofs. $F_{10}$ and $R_1$ are essential for answering the question and hence conditioning on the previously generated proofs will help the model adjust the relevance of nodes and edges in the subsequent proofs. Quantitatively, we find that about 75\% of the samples with 4 proofs have at least one node and one edge common across all the proofs (see Figure \ref{fig:cnce}). Thus, we propose \modelseq{} (see Figure \ref{fig:it-mprover}), which broadly consists of a base \textsc{PRover} architecture, as in Figure \ref{fig:ml-mprover} and an additional $p$ node and edge encoders for generating a maximum of $p$ proofs. The proofs are generated iteratively until an empty graph is generated to denote the end.

Base \textsc{PRover} architecture computes the first level of node embeddings $\textbf{N}^1 \in \mb{R}^{k \times d}$ and edge embeddings $\textbf{E}^1 \in \mb{R}^{k^2 \times d}$. These are passed respectively through a node and edge classifier to generate the node probabilities $np^1 \in \mb{R}^{k}$ and edge probabilities $ep^1 \in \mb{R}^{k^2}$, corresponding to the first proof. In the next iteration, two transformer encoders generate the node and edge embeddings corresponding to the second proof. Specifically, we condition the generation of the next node embeddings $\textbf{N}^2$ on the previous node ($\textbf{N}^1$) and edge ($\textbf{E}^1$) embeddings simultaneously. Conditioning on both is crucial because $\textbf{N}^1$ captures the relevance of nodes for the first proof, while $\textbf{E}^1$ contains information about the strength of the connections between these nodes. We condition $\textbf{E}^2$ only on $\textbf{E}^1$, because the edge embeddings corresponding to the nodes predicted by $\textbf{N}^1$ are already updated in $\textbf{E}^1$. Formally,
\begin{equation}
\small
\left.\begin{aligned}
    \textbf{T}^{1} &= W^{(1)} \textbf{E}^1 W^{(2)}, W^{(1)} \in \mb{R}^{k \times k^2},  W^{(2)} \in \mb{R}^{3d \times d} \nonumber \\
    \textbf{N}' &= [\textbf{N}^{1}; \textbf{T}^1]W^{(3)}, W^{(3)} \in \mb{R}^{2d \times d} \nonumber \\
    \textbf{N}^{2} &= \mathit{Transformer}(\textbf{N}') \nonumber;~~
    \textbf{E}^{2} = \mathit{Transformer}(\textbf{E}^{1}) \nonumber
\end{aligned}\right.
\end{equation}

These next set of embeddings, when passed through the respective node and edge classifiers, predict the node probabilities $np^2 \in \mb{R}^k$ and edge probabilities $ep^2 \in \mb{R}^{k^2}$, denoting the likelihood of their presence in the second proof.
We repeat this process of stacking up the node and edge encoders for generating a maximum of $p$ proofs. Given the node and edge probabilities corresponding to each proof and a QA probability from the QA module, we train \modelseq{} jointly with all proofs using the Hungarian Loss, described below.

\subsection{Permutation-Invariant Hungarian Loss}
\label{sec:hungarian}
Unlike words in text generation, proofs can be generated in any arbitrary order. Consequently, computing cross-entropy loss between the $i^{th}$ predicted proof and the $i^{th}$ gold proof, $i \in \{1, ..., p\}$ will be sub-optimal. Thus, we use a permutation-invariant Hungarian Loss \cite{Zhang2019deepsets, zhang2019fspool} which finds the most optimal assignment between the predicted proofs and the gold proofs such that the overall loss is minimized. Formally, the Hungarian loss $\mathcal{L}_H$ and total loss $\mathcal{L}$ are denoted as follows:
\begin{equation}
\small
\left.\begin{aligned}
  \mc{L}_H &= \min_{\pi \in \Pi} \sum_{i=1}^{p} \mathit{CE}(np^{i}, y_n^{\pi(i)})~+~\mathit{CE}(ep^{i}, y_e^{\pi(i)}) \nonumber \\
    \mc{L} &= \mc{L}_{QA} + \mc{L}_H \nonumber
\end{aligned}\right.
\end{equation}

where $\mathit{CE}(., .)$ is the cross entropy loss, $np^{i}$ and  $ep^{i}$ are the respective node and edge probabilities for the $i^{th}$ predicted proof while $y_n^{\pi(i)} \in \{0,1\}^{k}$ and $y_e^{\pi(i)} \in \{0,1\}^{k^2}$ are the respective true node and edge labels for the gold proof $\pi(i)$, where $\pi$ is the most optimal permutation. The Hungarian Loss is implemented by first summing the node and edge cross-entropy loss matrices $\mc{L}^n \in \mb{R}^{p \times p}$ and $\mc{L}^e \in \mb{R}^{p \times p}$ respectively, each entry $(i,j)$ of which corresponds to the proof loss between the $i^{th}$ predicted proof and $j^{th}$ gold proof (see Figures \ref{fig:ml-mprover} and \ref{fig:it-mprover}). Then we find the best assignment between the gold and predicted proofs through the Hungarian algorithm \cite{Kuhn55thehungarian}. Our final loss sums the Hungarian proof loss and the QA loss.

\begin{table*}[t]
\small
\centering
\begin{tabular}{lccccccccccc}
\toprule
           &    & \multicolumn{3}{c}{Node} & \multicolumn{3}{c}{Edge} & \multicolumn{3}{c}{Proof} \\ \cmidrule(lr){3-5} \cmidrule(lr){6-8} \cmidrule(lr){9-11}
           & QA & P      & R      & F1     & P      & R      & F1     & P    & R    & F1   & FA   \\ \midrule
\textsc{PRover} \cite{saha2020prover}     & 99.3 &   89.2 &	84.9 &	86.0 &	87.5 &	84.2 &	85.3 &	87.1 &	84.0 &	84.7 &	81.2       \\
\textsc{PRover}-all     & 99.3 &   87.9 &	83.8 &	84.9 &	87.1 &	83.6 &	84.6 &	85.9 &	82.8 &	83.7 &	80.3       \\
\textsc{PRover}-top-$p$ & 99.3 & 34.4 & 88.4 & 48.4 & 33.8 & 87.4 & 47.7 & 33.3 & 86.7 & 47.2 & 00.0  \\
\textsc{PRover}-top-$p$-classifier & 99.3 & 85.7 & 84.4 & 83.8 & 84.8 & 84.1 & 83.5 & 83.9 & 83.4 & 82.6 & 77.3     \\
\textsc{PRover}-top-$p$-threshold & 99.3 & 84.4 & 88.0 & 85.0 & 83.6 & 87.1 & 84.4 & 83.0 & 86.5 & 83.8 & 77.2     \\
\shortmodelmult{} & \textbf{99.5}    & 89.4       &  89.2      & 89.0       & 87.7     &  87.8      &  87.4      &  87.2    &  87.3    & 87.0     &  83.8    \\
\shortmodelseq{} & \textbf{99.5} & \textbf{90.6} & \textbf{90.2} & \textbf{90.0} & \textbf{89.6} & \textbf{89.4} & \textbf{89.2} & \textbf{89.1} & \textbf{89.0} & \textbf{88.7} & \textbf{85.5} \\
\bottomrule  
\end{tabular}
\vspace{-6pt}
\caption{\label{du5-test} Comparison of our \model{} models with \textsc{PRover} variations on DU5 test set. \modelseq{}'s improvement in Full Accuracy over \modelmult{} is statistically significant with $p < 0.001$.}
\vspace{-5pt}
\end{table*}

\vspace{-4pt}
\subsection{Integer Linear Program (ILP) Inference}
Following \textsc{PRover}, we generate valid proofs during inference using an ILP, subject to multiple global constraints (see \citet{saha2020prover}). For each predicted proof, the predicted nodes and edge
probabilities from \model{}, we obtain the 
corresponding predicted edges using Eqn.~\ref{eqn:edge_opt}.

\section{Experimental Setup}
We experiment on synthetic, hand-authored zero-shot, and human paraphrased datasets, following \citet{clark2020transformers,saha2020prover}.

\textbf{Datasets:} The five synthetic datasets \textbf{DU0-DU5} consist of 100k questions with their own train, validation and test splits (70/10/20) and reasoning depths up to $D=0,1,2,3,5$. Each example in these datasets is annotated with all possible proofs. The second dataset is a \textbf{Birds-Electricity} dataset, consisting of 5k hand-authored samples aimed at evaluating the zero-shot performance of the models. Unlike the previous datasets, all examples in this dataset have a unique gold proof. Third, \textbf{ParaRules} is a human-paraphrased dataset, consisting of 40k examples with all possible proofs, where the facts and rules are paraphrased by crowdworkers. Further details of the datasets and model's hyperparameters can be found in the appendix.

\textbf{Evaluation Metrics: } Following \textsc{PRover}, QA evaluation is done through accuracy. For proofs, we compute the following metrics: (1) \textbf{Node Precision, Recall, F1} (2) \textbf{Edge Precision, Recall, F1}, (3) \textbf{Proof Precision, Recall, F1}, and (4) \textbf{Full Accuracy (FA)}. For each sample, given a set of gold proofs and predicted proofs, node precision is computed as the fraction of predicted proofs where the predicted node set matches exactly with a gold proof's node set. Similarly, node recall for each sample is computed as the fraction of gold proofs where the corresponding node sets match exactly. The overall node precision, recall, and F1 are the respective sample-wise precision, recall, and F1 scores averaged over all the samples. Edge metrics are computed similarly but with respect to the edges only and the proof metrics consider both nodes and edges in conjunction. Our final metric, full accuracy evaluates a sample as a whole and is given by the fraction of samples where the answer and all corresponding proofs are exactly correct.

\section{Results and Analysis}

\subsection{Comparison of \textsc{PRover} variants with \model{}}
\label{exp:main}
In Table \ref{du5-test}, we compare \shortmodelmult{} and \shortmodelseq{} with five variants of \textsc{PRover} -- (1) \textsc{PRover}, as introduced in \citet{saha2020prover}, trained with one proof per example and also generates a single proof, (2) \textsc{PRover}-all, trained with all proofs as separate examples and generates a single proof per example, (3) \textsc{PRover}-top-$p$, an extension of \textsc{PRover}-all, generating top-$p$ proofs for all examples, (4) \textsc{PRover}-top-$p$-classifier, an improvement over the vanilla top-$p$ model, where we first predict the number of proofs by training a RoBERTa classifier with concatenated question and context and then generate those many top proof graphs, and (5) \textsc{PRover}-top-$p$-threshold, another improved model over vanilla top-$p$, where we use the optimization score from Equation \ref{eqn:node_inferenc} to predict the number of proofs to generate, i.e., we stop generating proofs when the score difference between two consecutive proofs exceeds a certain threshold (tuned on the validation set). All models are trained on the DU5 train set and tested on the corresponding test set. Based on Figure \ref{fig:percent_proof} which shows that 98\% of the dataset contains samples with $\leq$ 3 proofs, we set max-proofs, $p = 3$. 87\% of the examples in the dataset have a single gold proof, thereby making \textsc{PRover} a strong baseline.

\begin{table*}
\small
\centering
\begin{tabular}{lccccccccccc}
\toprule
           &    & \multicolumn{3}{c}{Node} & \multicolumn{3}{c}{Edge} & \multicolumn{4}{c}{Proof} \\ \cmidrule(lr){3-5} \cmidrule(lr){6-8} \cmidrule(lr){9-11}
           & QA & P      & R      & F1     & P      & R      & F1     & P    & R    & F1   & FA   \\ \midrule
\textsc{PRover} \cite{saha2020prover}     & \textbf{86.5} &	81.3 &	81.3 &	81.3 &	81.4 &	81.4 &	81.4 &	80.7 &	80.7 &	80.7 &	80.7       \\
\textsc{PRover}-all  & 85.9 &  80.9 & 80.9 & 80.9 & 80.4 & 80.4 & 80.4 & 80.2 & 80.2 & 80.2 & 80.0     \\
\shortmodelmult{}  & 85.1 &  79.2 & 79.9 & 79.4 & 79.4 & 79.9 & 79.5 & 78.7 & 79.1 & 78.8 & 78.1 \\
\shortmodelseq{} & 86.3 & \textbf{82.7} & \textbf{83.3} & \textbf{82.9} & \textbf{82.4} & \textbf{83.0} & \textbf{82.6} & \textbf{82.2} & \textbf{82.7} & \textbf{82.3} & \textbf{81.8} \\ 
\bottomrule  
\end{tabular}
\vspace{-5pt}
\caption{\label{birds-elec} Comparison of all models on the zero-shot Birds-Electricity dataset containing one gold proof per sample. \modelseq{}'s improvement in Full Accuracy over \textsc{PRover} is statistically significant with $p < 0.001$.}
\vspace{-5pt}
\end{table*}

We observe that \textsc{PRover}-all has a slightly lower proof F1 than \textsc{PRover}, because the model likely gets confused with multiple possible proofs for the same context and question. \textsc{PRover}-top-$p$'s huge drop in precision is unsurprising because the subsequent non-empty proofs are always incorrect, causing full accuracy to drop to 0\%. When we perform careful inference over \textsc{PRover} either by predicting the number of proofs or by thresholding and do not generate a fixed $p$ number of proofs for all examples, we observe a boost in precision over the vanilla top-$p$ model, with very little drop in recall. However, \textsc{PRover} continues to be a stronger baseline than all the top-$p$ variants because of a lot of single-proof examples in the dataset.

Both \model{} models improve significantly on the state-of-the-art proof F1, while retaining a near perfect QA accuracy. \shortmodelseq{} is a significantly stronger model because of its explicit conditioning mechanism and obtains up to a statistically significant\footnote{We use bootstrap test \cite{efron1994introduction} for calculating the statistical significance score.} ($p < 0.001$) 4\% improvement on proof F1 and full accuracy. While our model is expected to improve the proof recall compared to \textsc{PRover} and \textsc{PRover}-all because of the generation of multiple proofs, the improvement in precision is particularly important as it shows that the subsequently generated proofs by \shortmodelseq{} are mostly correct. Similarly, its improvement in proof recall compared to \textsc{PRover}-top-$p$ also shows the strength of the model considering that \textsc{PRover}-top-$p$ generates the maximum number of proofs for every sample. Overall, \shortmodelseq{} outperforms all other models in all metrics. In summary, careful inference strategies over a single-proof generation model like \textsc{PRover} are largely ineffective for generating multiple proofs and an effective proof-set generation model needs to exploit and learn the inter-proof correlations during the training phase itself. Our experiments on the ParaRules dataset demonstrate similar findings, details of which and the effect of varying $p$ for \model{} is in the appendix. 

\modelseq{} performs equally well on the subset of questions where the context has \emph{negations}, achieving a high proof F1 of $90.8$. As part of error analysis, we find that 58\% of \modelseq{}'s wrongly predicted proofs have more nodes and edges than those in the gold proof, suggesting that our model tends to overestimate the essential rules and facts and their inter-connections. In the following subsections, we analyze \model{}'s generalization capabilities in three different contexts -- zero-shot settings, higher depth questions, and training with less training data.

\begin{table}[t]
\small
\centering
\begin{tabular}{cccccccc}
\toprule
          & & \multicolumn{3}{c}{Proof F1} & \multicolumn{3}{c}{Full Acc} \\ \cmidrule(lr){3-5} \cmidrule(lr){6-8}
         d  & MP & PR & ML      & IT & PR & ML & IT   \\ \midrule
0  & 7.2 & 93.8 & 97.8 & \textbf{98.2} & 92.6 & 96.7 & \textbf{97.0}    \\
1  & 10.3 & 88.0 & 92.8 & \textbf{93.5} & 85.7 & 91.0 & \textbf{91.7}\\ 
2 & 15.7 & 80.8 & 86.1 & \textbf{87.1} & 76.5 & 81.8 & \textbf{83.7}\\
3 & 17.7 & 78.0 & 80.7 & \textbf{83.0} & 72.2 & 75.9 & \textbf{78.0} \\
4 & 19.9 & 71.1 & 72.3 & \textbf{77.2} & 65.9 & 66.4 & \textbf{70.1}\\
5 & 23.1 & 67.7 & 64.9 & \textbf{70.6} & 61.0 & 58.7 & \textbf{63.7}\\

\bottomrule  
\end{tabular}
\vspace{-5pt}
\caption{\label{du5-depth} Comparison of \textsc{PRover}-all and \model{} models on the subset of samples in DU5 test set requiring d depth of reasoning.}
\vspace{-10pt}
\end{table}

\subsection{Generalization to Zero-Shot Dataset with Single Gold Proofs}

The Birds-Electricity test-only dataset evaluates the zero-shot performance. It contains examples with single gold proofs; hence, if a multiple-proof generation model like \model{} transfers well to it, this indicates strong generalization capabilities because along with generating correct proofs, it also needs to infer the correct number of proofs. With that motivation, in Table \ref{birds-elec}, we compare \textsc{PRover} and \textsc{PRover}-all, both trained on DU5 to generate a single proof, with our \model{} models, also trained on DU5 and find that \shortmodelseq{} obtains state-of-the-art result on all proof-related metrics, while retaining the QA performance. Note that \shortmodelseq{} has two important design choices which explain its good performance on out-of-domain transfer -- (1) it trains on all proofs jointly, (2) explicit proof conditioning. Both of these, when combined, enable it to learn the correlations between the proofs to identify the degree of relevance of facts and rules, ranging from essential to sometimes useful to irrelevant, for a given question. Thus, on out-of-domain test data, it assigns soft prior relevance scores to the context which helps it to better learn the significantly smaller space of correct proofs and be more accurate even for a single-proof dataset.

\subsection{Generalization to Higher Depths}
The DU5 dataset consists of questions requiring reasoning up to a maximum depth of 5. Thus, we test the generalization capabilities of the \model{} models on higher depth questions. Specifically, in Table \ref{du5-depth}, we compare the DU5-trained models of \textsc{PRover}-all, \shortmodelmult{} and \shortmodelseq{} on the subset of DU5 test examples with varying depths of reasoning ($d$). Each row also shows the percentage of examples with multiple gold proofs (MP) which, unsurprisingly, increases as the depth increases. We observe that much of \shortmodelseq{}'s improvement compared to \shortmodelmult{} comes at higher depths where the presence of multiple proofs is a more frequent phenomenon. At depth-5, where 23\% of the examples have $>$ 1 correct proof, \shortmodelseq{} obtains a 6\% improvement over \shortmodelmult{}. This shows that joint training with all proofs and explicit conditioning between them leads to better generalization at higher depths.

\begin{table}
\small
\centering
\begin{tabular}{lccccccc}
\toprule
          & \multicolumn{2}{c}{QA}  & \multicolumn{2}{c}{Proof F1} & \multicolumn{2}{c}{Full Acc} \\ \cmidrule(lr){2-3} \cmidrule(lr){4-5} \cmidrule(lr){6-7}
         Count & ML & IT & ML    & IT   & ML & IT   \\ \midrule
         
         10k & \textbf{87.2} & 86.1 & \textbf{41.5} & 41.4 & 39.0 & \textbf{39.5}\\ 
         30k & 97.7 & \textbf{98.2} & 74.3 & \textbf{74.9} & 71.2 & \textbf{72.0}\\
         50k & \textbf{99.4} & \textbf{99.4} & 83.7 & \textbf{84.5} & 80.0 & \textbf{81.0}\\
         70k (All) & \textbf{99.5} & \textbf{99.5} & 87.0 & \textbf{88.7} & 83.8 & \textbf{85.5}\\

\bottomrule 
\end{tabular}
\vspace{-5pt}
\caption{\label{less-data} Comparative study between the two \model{} models with varying amount of training data on DU5. Count = number of training examples.}
\vspace{-3pt}
\end{table}

\begin{figure*}
    \centering
    \includegraphics[width=0.99\textwidth]{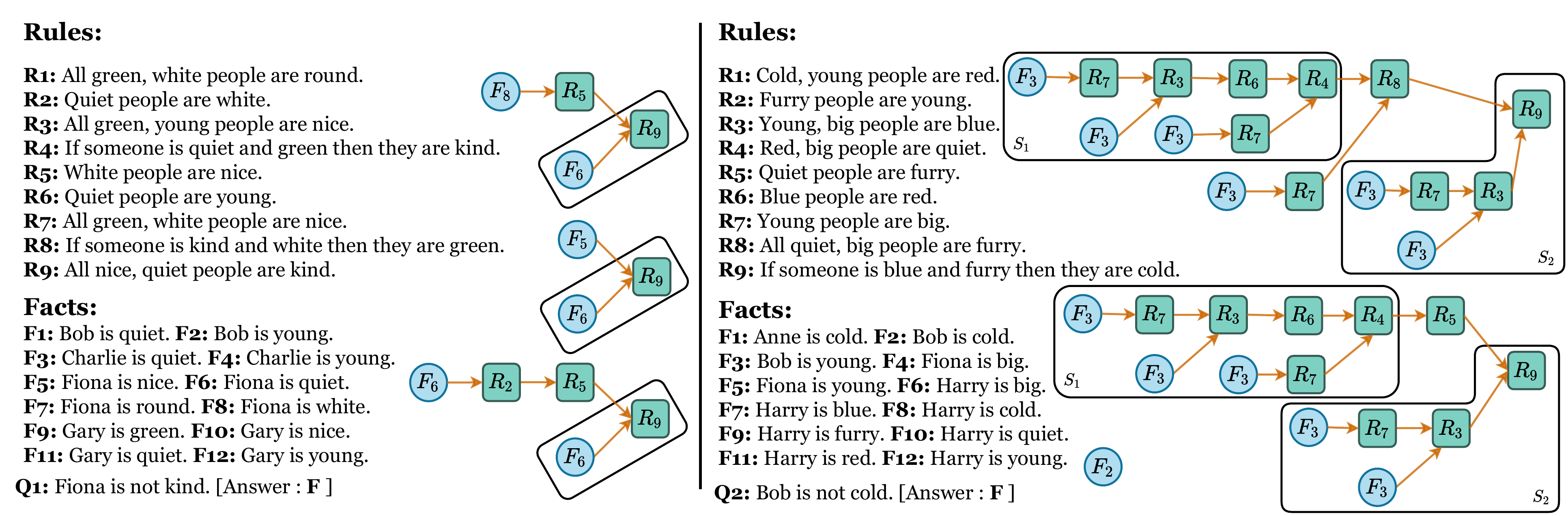}
    \vspace{-5pt}
    \caption{All proofs correctly generated by our \modelseq{} model for two randomly chosen questions corresponding to two different rule-bases.\vspace{-5pt}}
    \vspace{-8pt}
    \label{fig:example2}
\end{figure*}

\subsection{Generalization with Less Training Data}

Collecting proofs for supervised training is expensive in most real-world scenarios. Hence, on top of the zero-shot and depth generalization results presented so far, we ask if our \model{} models can learn from less training data. Table \ref{less-data} shows that these models obtain near perfect QA accuracy with only 40\% of the training data (30k examples). However, proof generation proves to be challenging and only improves with sufficient training data. Another interesting observation is that while both \model{} models perform comparably with less training data, \shortmodelseq{} starts to outperform \shortmodelmult{} upon training with more examples. \shortmodelseq{} consists of more trainable parameters because of its multiple node and edge encoders, which get learned better with more data. See appendix for runtime and parameter space of these models.

\subsection{Comparison of \model{} with the Skyline Single-Proof Generation Model}
\begin{figure}
\centering
    \includegraphics[width=0.95\columnwidth]{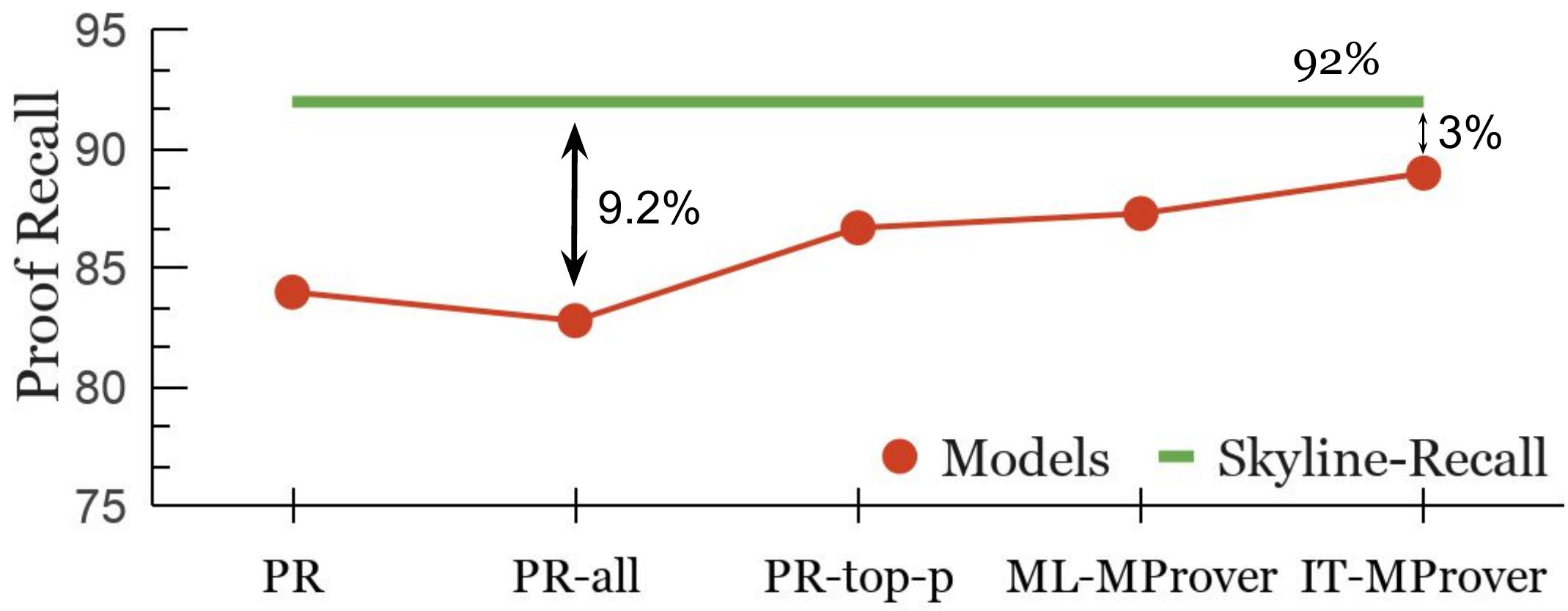}
    \vspace{-5pt}
\caption{Comparison of proof recall for all models with that of the skyline single-proof generation model.}
\label{fig:proof_acc_graph}
\vspace{-12pt}
\end{figure}

We find that an ideal (skyline) single-proof generation model's proof recall for the DU5 dataset is upper-bounded by 92\% as it contains about 87\% of single-proof examples. This is computed by considering exactly 1 correct proof per question. Hence, we ask how well our \model{} models compare with this ideal performance (Figure \ref{fig:proof_acc_graph}). Our results are encouraging, not only because \shortmodelseq{} generates more correct proofs than all other models but also because it almost matches the performance of the skyline single-proof generation model. 
The \textsc{PRover} model is $9.2 \%$ worse as compared to the skyline single-proof generation model while \shortmodelseq{} reduces this gap to $3\%$. Given the dataset mostly contains single-proof examples, the skyline is a strong upper-bound on proof generation performance and \shortmodelseq{} significantly reduces the gap. See appendix for ablations of \shortmodelseq{}, including the effect of Hungarian Loss.

\section{Qualitative Analysis of \model{}}

Fig.~\ref{fig:example2} shows the sets of proofs correctly generated by \modelseq{} for two randomly chosen questions. For $Q_1$, it generates all the possible proofs by identifying the common subgraph ${F_6 \rightarrow R_9}$. $Q_2$ is interesting, because (i) the single-node proof $F_2$ is significantly different from the other proofs in both structure and size, and (ii) the two larger proofs have two distinct common subgraphs. Here, \textsc{PRover} performs a simple lookup in the rule-base to generate the proof $F_2$, thereby limiting our understanding of its reasoning capabilities. However,  \model{}, through its ability to also generate the larger and more complex proofs enhances the transparency and verification of its reasoning abilities, and hence is a crucial step towards bridging the gap between neural and symbolic approaches.

\section{Conclusion} 
\vspace{-3pt}
We proposed \modelmult{} and \modelseq{}, two variants of a proof-set generation model where the former performs implicit conditioning between the proofs to generate them in parallel while the latter generates a proof-set through explicit conditioning on the previously generated proofs.
Both models obtain strong proof F1 improvements on synthetic and human-paraphrased datasets and \modelseq{} also obtains state-of-the-art proof F1 on a zero-shot dataset with single proofs. \model{}'s modeling is fairly generic and similar methods can be used in generating a set of structured explanations for other NLP tasks like multi-hop QA. 

\section*{Ethical Considerations}
Despite the overwhelming success of pre-trained language models for various NLP tasks, a common criticism is their lack of interpretability. Generating structured proofs from such models allows us to explain their reasoning capabilities and also bridges the gap between neural and symbolic systems. In this work, we take a step closer towards improving the interpretability of rule-based reasoning by generating a set of multiple proofs, each providing a diverse rationale for the reasoning process. We experiment with a wide variety of rule-bases ranging from synthetic to hand-authored to human-paraphrased rule-bases. Our results show good generalization performance of our models across three different aspects -- (1) zero-shot settings, (2) questions requiring higher depths of reasoning, and (3) availability of less training data. We hope our models and findings will inspire future work on generating multiple structured explanations for different compositional reasoning tasks in NLP.

\section*{Acknowledgements}
We thank the reviewers and Peter Hase for their helpful feedback. This work was supported by DARPA MCS
Grant N66001-19-2-4031, NSF-CAREER Award 1846185, DARPA YFA17-D17AP00022, ONR Grant N00014-18-1-2871, Microsoft Investigator Fellowship, and Munroe \& Rebecca Cobey Fellowship. The views in this article are those of the
authors and not the funding agency.

\bibliography{naacl2021}

\begin{thebibliography}{57}
\expandafter\ifx\csname natexlab\endcsname\relax\def\natexlab#1{#1}\fi

\bibitem[{Bahdanau et~al.(2015)Bahdanau, Cho, and Bengio}]{bahdanau2014neural}
Dzmitry Bahdanau, Kyunghyun Cho, and Yoshua Bengio. 2015.
\newblock Neural machine translation by jointly learning to align and
  translate.
\newblock In \emph{ICLR}.

\bibitem[{Berant et~al.(2013)Berant, Chou, Frostig, and
  Liang}]{berant2013semantic}
Jonathan Berant, Andrew Chou, Roy Frostig, and Percy Liang. 2013.
\newblock \href {https://www.aclweb.org/anthology/D13-1160.pdf} {Semantic
  parsing on freebase from question-answer pairs}.
\newblock In \emph{Proceedings of the 2013 conference on empirical methods in
  natural language processing}, pages 1533--1544.

\bibitem[{Berant and Liang(2014)}]{berant2014semantic}
Jonathan Berant and Percy Liang. 2014.
\newblock \href {https://www.aclweb.org/anthology/P14-1133} {Semantic parsing
  via paraphrasing}.
\newblock In \emph{Proceedings of the 52nd Annual Meeting of the Association
  for Computational Linguistics (Volume 1: Long Papers)}, pages 1415--1425.

\bibitem[{Camburu et~al.(2018)Camburu, Rockt{\"a}schel, Lukasiewicz, and
  Blunsom}]{camburu2018snli}
Oana-Maria Camburu, Tim Rockt{\"a}schel, Thomas Lukasiewicz, and Phil Blunsom.
  2018.
\newblock \href
  {http://papers.nips.cc/paper/8163-e-snli-natural-language-inference-with-natural-language-explanations.pdf}
  {e-{SNLI}: Natural language inference with natural language explanations}.
\newblock In \emph{Advances in Neural Information Processing Systems}, pages
  9539--9549.

\bibitem[{Cho et~al.(2014)Cho, van Merrienboer, G{\"{u}}l{\c{c}}ehre, Bougares,
  Schwenk, and Bengio}]{cho2014mt}
Kyunghyun Cho, Bart van Merrienboer, {\c{C}}aglar G{\"{u}}l{\c{c}}ehre, Fethi
  Bougares, Holger Schwenk, and Yoshua Bengio. 2014.
\newblock \href {http://arxiv.org/abs/1406.1078} {Learning phrase
  representations using {RNN} encoder-decoder for statistical machine
  translation}.
\newblock In \emph{EMNLP}.

\bibitem[{Clark et~al.(2020)Clark, Tafjord, and
  Richardson}]{clark2020transformers}
Peter Clark, Oyvind Tafjord, and Kyle Richardson. 2020.
\newblock \href {https://doi.org/10.24963/ijcai.2020/537} {Transformers as soft
  reasoners over language}.
\newblock In \emph{Proceedings of the Twenty-Ninth International Joint
  Conference on Artificial Intelligence, {IJCAI-20}}, pages 3882--3890.
  International Joint Conferences on Artificial Intelligence Organization.
\newblock Main track.

\bibitem[{Dai et~al.(2017)Dai, Fidler, Urtasun, and Lin}]{dai2017towards}
Bo~Dai, Sanja Fidler, Raquel Urtasun, and Dahua Lin. 2017.
\newblock Towards diverse and natural image descriptions via a conditional gan.
\newblock In \emph{Proceedings of the IEEE International Conference on Computer
  Vision}, pages 2970--2979.

\bibitem[{DeYoung et~al.(2020)DeYoung, Jain, Rajani, Lehman, Xiong, Socher, and
  Wallace}]{deyoung2020eraser}
Jay DeYoung, Sarthak Jain, Nazneen~Fatema Rajani, Eric Lehman, Caiming Xiong,
  Richard Socher, and Byron~C Wallace. 2020.
\newblock Eraser: A benchmark to evaluate rationalized nlp models.
\newblock In \emph{Proceedings of the 58th Annual Meeting of the Association
  for Computational Linguistics}, pages 4443--4458.

\bibitem[{Doshi-Velez and Kim(2017)}]{doshi2017towards}
Finale Doshi-Velez and Been Kim. 2017.
\newblock \href {https://arxiv.org/abs/1702.08608} {Towards a rigorous science
  of interpretable machine learning}.
\newblock \emph{arXiv preprint arXiv:1702.08608}.

\bibitem[{Efron and Tibshirani(1994)}]{efron1994introduction}
Bradley Efron and Robert~J Tibshirani. 1994.
\newblock \emph{An introduction to the bootstrap}.
\newblock CRC press.

\bibitem[{Gao et~al.(2019)Gao, Chen, Chenthamarakshan, and
  Witbrock}]{gao2019sequential}
Tian Gao, Jie Chen, Vijil Chenthamarakshan, and Michael Witbrock. 2019.
\newblock A sequential set generation method for predicting set-valued outputs.
\newblock In \emph{Proceedings of the AAAI Conference on Artificial
  Intelligence}, volume~33, pages 2835--2842.

\bibitem[{Gimpel et~al.(2013)Gimpel, Batra, Dyer, and
  Shakhnarovich}]{gimpel2013}
Kevin Gimpel, Dhruv Batra, Chris Dyer, and Gregory Shakhnarovich. 2013.
\newblock \href {https://www.aclweb.org/anthology/D13-1111} {A systematic
  exploration of diversity in machine translation}.
\newblock In \emph{Proceedings of the 2013 Conference on Empirical Methods in
  Natural Language Processing}, pages 1100--1111, Seattle, Washington, USA.
  Association for Computational Linguistics.

\bibitem[{Gontier et~al.(2020)Gontier, Sinha, Reddy, and
  Pal}]{gontier2020measuring}
Nicolas Gontier, Koustuv Sinha, Siva Reddy, and Chris Pal. 2020.
\newblock Measuring systematic generalization in neural proof generation with
  transformers.
\newblock \emph{Advances in Neural Information Processing Systems}, 33.

\bibitem[{Hase and Bansal(2020)}]{hase_evaluating_2020}
Peter Hase and Mohit Bansal. 2020.
\newblock \href {https://arxiv.org/abs/2005.01831} {Evaluating explainable
  {AI}: Which algorithmic explanations help users predict model behavior?}
\newblock In \emph{Proceedings of the 58th Annual Meeting of the Association
  for Computational Linguistics}.

\bibitem[{Jacovi and Goldberg(2020)}]{jacovi2020towards}
Alon Jacovi and Yoav Goldberg. 2020.
\newblock Towards faithfully interpretable nlp systems: How should we define
  and evaluate faithfulness?
\newblock In \emph{ACL}.

\bibitem[{Jain et~al.(2017)Jain, Zhang, and Schwing}]{jain2017creativity}
Unnat Jain, Ziyu Zhang, and Alexander~G Schwing. 2017.
\newblock Creativity: Generating diverse questions using variational
  autoencoders.
\newblock In \emph{Proceedings of the IEEE Conference on Computer Vision and
  Pattern Recognition}, pages 6485--6494.

\bibitem[{Jansen and Ustalov(2019)}]{jansen2019textgraphs}
Peter Jansen and Dmitry Ustalov. 2019.
\newblock Textgraphs 2019 shared task on multi-hop inference for explanation
  regeneration.
\newblock In \emph{Proceedings of the Thirteenth Workshop on Graph-Based
  Methods for Natural Language Processing (TextGraphs-13)}, pages 63--77.

\bibitem[{Jansen et~al.(2018)Jansen, Wainwright, Marmorstein, and
  Morrison}]{jansen2018worldtree}
Peter~A Jansen, Elizabeth Wainwright, Steven Marmorstein, and Clayton~T
  Morrison. 2018.
\newblock Worldtree: A corpus of explanation graphs for elementary science
  questions supporting multi-hop inference.
\newblock \emph{arXiv preprint arXiv:1802.03052}.

\bibitem[{Jhamtani and Clark(2020)}]{jhamtani2020learning}
Harsh Jhamtani and Peter Clark. 2020.
\newblock Learning to explain: Datasets and models for identifying valid
  reasoning chains in multihop question-answering.
\newblock In \emph{Proceedings of the 2020 Conference on Empirical Methods in
  Natural Language Processing (EMNLP)}, pages 137--150.

\bibitem[{Khot et~al.(2020)Khot, Clark, Guerquin, Jansen, and
  Sabharwal}]{khot2020qasc}
Tushar Khot, Peter Clark, Michal Guerquin, Peter Jansen, and Ashish Sabharwal.
  2020.
\newblock Qasc: A dataset for question answering via sentence composition.
\newblock In \emph{Proceedings of the AAAI Conference on Artificial
  Intelligence}, 05, pages 8082--8090.

\bibitem[{Kosiorek et~al.(2020)Kosiorek, Kim, and
  Rezende}]{kosiorek2020conditional}
Adam~R Kosiorek, Hyunjik Kim, and Danilo~J Rezende. 2020.
\newblock Conditional set generation with transformers.
\newblock \emph{arXiv preprint arXiv:2006.16841}.

\bibitem[{Kuhn and Yaw(1955)}]{Kuhn55thehungarian}
H.~W. Kuhn and Bryn Yaw. 1955.
\newblock The hungarian method for the assignment problem.
\newblock \emph{Naval Res. Logist. Quart}, pages 83--97.

\bibitem[{Kulikov et~al.(2018)Kulikov, Miller, Cho, and
  Weston}]{kulikov2018importance}
Ilya Kulikov, Alexander~H Miller, Kyunghyun Cho, and Jason Weston. 2018.
\newblock Importance of a search strategy in neural dialogue modelling.
\newblock In \emph{iNLG}.

\bibitem[{Kumar and Talukdar(2020)}]{kumar2020nile}
Sawan Kumar and Partha Talukdar. 2020.
\newblock {NILE}: Natural language inference with faithful natural language
  explanations.
\newblock In \emph{ACL}.

\bibitem[{Lee et~al.(2018)Lee, Lee, Kim, Kosiorek, Choi, and
  Teh}]{lee2018settoset}
Juho Lee, Yoonho Lee, Jungtaek Kim, Adam~R. Kosiorek, Seungjin Choi, and
  Yee~Whye Teh. 2018.
\newblock Set transformer: A framework for attention-based
  permutation-invariant neural networks.
\newblock In \emph{ICML}.

\bibitem[{Lei et~al.(2016)Lei, Barzilay, and Jaakkola}]{lei2016rationalizing}
Tao Lei, Regina Barzilay, and Tommi Jaakkola. 2016.
\newblock Rationalizing neural predictions.
\newblock In \emph{Proceedings of the 2016 Conference on Empirical Methods in
  Natural Language Processing}, pages 107--117.

\bibitem[{Li et~al.(2016)Li, Monroe, and Jurafsky}]{li2016simple}
Jiwei Li, Will Monroe, and Dan Jurafsky. 2016.
\newblock A simple, fast diverse decoding algorithm for neural generation.
\newblock \emph{arXiv preprint arXiv:1611.08562}.

\bibitem[{Lin et~al.(2019)Lin, Tafjord, Clark, and
  Gardner}]{lin-etal-2019-reasoning}
Kevin Lin, Oyvind Tafjord, Peter Clark, and Matt Gardner. 2019.
\newblock \href {https://doi.org/10.18653/v1/D19-5808} {Reasoning over
  paragraph effects in situations}.
\newblock In \emph{Proceedings of the 2nd Workshop on Machine Reading for
  Question Answering}, pages 58--62, Hong Kong, China. Association for
  Computational Linguistics.

\bibitem[{Liu et~al.(2019)Liu, Ott, Goyal, Du, Joshi, Chen, Levy, Lewis,
  Zettlemoyer, and Stoyanov}]{liu2019roberta}
Yinhan Liu, Myle Ott, Naman Goyal, Jingfei Du, Mandar Joshi, Danqi Chen, Omer
  Levy, Mike Lewis, Luke Zettlemoyer, and Veselin Stoyanov. 2019.
\newblock \href {https://arxiv.org/abs/1907.11692} {Ro{BERT}a: A robustly
  optimized bert pretraining approach}.
\newblock \emph{arXiv preprint arXiv:1907.11692}.

\bibitem[{MacCartney and Manning(2014)}]{maccartney2014natural}
Bill MacCartney and Christopher~D Manning. 2014.
\newblock \href {https://link.springer.com/chapter/10.1007/978-94-007-7284-7_8}
  {Natural logic and natural language inference}.
\newblock In \emph{Computing meaning}, pages 129--147. Springer.

\bibitem[{Musen and Van Der~Lei(1988)}]{musen1988brittleness}
Mark~A Musen and Johan Van Der~Lei. 1988.
\newblock \href
  {https://www.sciencedirect.com/science/article/pii/B9780444871374500291} {Of
  brittleness and bottlenecks: Challenges in the creation of
  pattern-recognition and expert-system models}.
\newblock In \emph{Machine Intelligence and Pattern Recognition}, volume~7,
  pages 335--352. Elsevier.

\bibitem[{Newell and Simon(1956)}]{newell1956logic}
Allen Newell and Herbert Simon. 1956.
\newblock \href
  {https://ieeexplore.ieee.org/stamp/stamp.jsp?tp=&arnumber=1056797} {The logic
  theory machine--a complex information processing system}.
\newblock \emph{IRE Transactions on information theory}, 2(3):61--79.

\bibitem[{Prasad et~al.(2014)Prasad, Jegelka, and Batra}]{prasad2014}
Adarsh Prasad, Stefanie Jegelka, and Dhruv Batra. 2014.
\newblock \href
  {https://proceedings.neurips.cc/paper/2014/file/8d9a0adb7c204239c9635426f35c9522-Paper.pdf}
  {Submodular meets structured: Finding diverse subsets in exponentially-large
  structured item sets}.
\newblock In \emph{Advances in Neural Information Processing Systems},
  volume~27, pages 2645--2653. Curran Associates, Inc.

\bibitem[{Raffel et~al.(2020)Raffel, Shazeer, Roberts, Lee, Narang, Matena,
  Zhou, Li, and Liu}]{raffel2019t5}
Colin Raffel, Noam Shazeer, Adam Roberts, Katherine Lee, Sharan Narang, Michael
  Matena, Yanqi Zhou, Wei Li, and Peter~J. Liu. 2020.
\newblock \href {http://jmlr.org/papers/v21/20-074.html} {Exploring the limits
  of transfer learning with a unified text-to-text transformer}.
\newblock \emph{Journal of Machine Learning Research}, 21(140):1--67.

\bibitem[{Rajani et~al.(2019)Rajani, McCann, Xiong, and
  Socher}]{rajani2019explain}
Nazneen~Fatema Rajani, Bryan McCann, Caiming Xiong, and Richard Socher. 2019.
\newblock \href {https://www.aclweb.org/anthology/P19-1487.pdf} {Explain
  yourself! leveraging language models for commonsense reasoning}.
\newblock In \emph{Proceedings of the 57th Annual Meeting of the Association
  for Computational Linguistics}, pages 4932--4942.

\bibitem[{Richardson et~al.(2020)Richardson, Hu, Moss, and
  Sabharwal}]{richardson2020probing}
Kyle Richardson, Hai Hu, Lawrence Moss, and Ashish Sabharwal. 2020.
\newblock Probing natural language inference models through semantic fragments.
\newblock In \emph{Proceedings of the AAAI Conference on Artificial
  Intelligence}, volume~34, pages 8713--8721.

\bibitem[{Rudin(2019)}]{rudin2019stop}
Cynthia Rudin. 2019.
\newblock \href {https://www.nature.com/articles/s42256-019-0048-x} {Stop
  explaining black box machine learning models for high stakes decisions and
  use interpretable models instead}.
\newblock \emph{Nature Machine Intelligence}, 1(5):206--215.

\bibitem[{Saha et~al.(2020)Saha, Ghosh, Srivastava, and
  Bansal}]{saha2020prover}
Swarnadeep Saha, Sayan Ghosh, Shashank Srivastava, and Mohit Bansal. 2020.
\newblock {PR}over: Proof generation for interpretable reasoning over rules.
\newblock In \emph{Proceedings of the 2020 Conference on Empirical Methods in
  Natural Language Processing (EMNLP)}, pages 122--136.

\bibitem[{Shen et~al.(2019)Shen, Ott, Auli, and Ranzato}]{shen2019mixture}
Tianxiao Shen, Myle Ott, Michael Auli, and Marc'Aurelio Ranzato. 2019.
\newblock \href {http://proceedings.mlr.press/v97/shen19c.html} {Mixture models
  for diverse machine translation: Tricks of the trade}.
\newblock In \emph{Proceedings of the 36th International Conference on Machine
  Learning}, volume~97 of \emph{Proceedings of Machine Learning Research},
  pages 5719--5728, Long Beach, California, USA. PMLR.

\bibitem[{Shu et~al.(2019)Shu, Nakayama, and Cho}]{shu2019generating}
Raphael Shu, Hideki Nakayama, and Kyunghyun Cho. 2019.
\newblock Generating diverse translations with sentence codes.
\newblock In \emph{Proceedings of the 57th Annual Meeting of the Association
  for Computational Linguistics}, pages 1823--1827.

\bibitem[{Tafjord et~al.(2019)Tafjord, Gardner, Lin, and
  Clark}]{tafjord-etal-2019-quartz}
Oyvind Tafjord, Matt Gardner, Kevin Lin, and Peter Clark. 2019.
\newblock \href {https://doi.org/10.18653/v1/D19-1608} {{Q}ua{RT}z: An
  open-domain dataset of qualitative relationship questions}.
\newblock In \emph{Proceedings of the 2019 Conference on Empirical Methods in
  Natural Language Processing and the 9th International Joint Conference on
  Natural Language Processing (EMNLP-IJCNLP)}, pages 5941--5946, Hong Kong,
  China. Association for Computational Linguistics.

\bibitem[{Vaswani et~al.(2017)Vaswani, Shazeer, Parmar, Uszkoreit, Jones,
  Gomez, Kaiser, and Polosukhin}]{vaswani2017attention}
Ashish Vaswani, Noam Shazeer, Niki Parmar, Jakob Uszkoreit, Llion Jones,
  Aidan~N Gomez, {\L}ukasz Kaiser, and Illia Polosukhin. 2017.
\newblock \href
  {http://papers.nips.cc/paper/7181-attention-is-all-you-need.pdf} {Attention
  is all you need}.
\newblock In \emph{Advances in neural information processing systems}, pages
  5998--6008.

\bibitem[{Vijayakumar et~al.(2018)Vijayakumar, Cogswell, Selvaraju, Sun, Lee,
  Crandall, and Batra}]{vijayakumar2016diversebeam}
Ashwin~K. Vijayakumar, Michael Cogswell, Ramprasaath~R. Selvaraju, Qing Sun,
  Stefan Lee, David~J. Crandall, and Dhruv Batra. 2018.
\newblock \href {http://arxiv.org/abs/1610.02424} {Diverse beam search:
  Decoding diverse solutions from neural sequence models}.
\newblock In \emph{{Proceedings of the AAAI Conference on Artificial
  Intelligence}}.

\bibitem[{{Vinyals} et~al.(2015){Vinyals}, {Toshev}, {Bengio}, and
  {Erhan}}]{vinyals2014caption}
O.~{Vinyals}, A.~{Toshev}, S.~{Bengio}, and D.~{Erhan}. 2015.
\newblock \href {https://doi.org/10.1109/CVPR.2015.7298935} {Show and tell: A
  neural image caption generator}.
\newblock In \emph{2015 IEEE Conference on Computer Vision and Pattern
  Recognition (CVPR)}, pages 3156--3164.

\bibitem[{Weston et~al.(2015)Weston, Bordes, Chopra, Rush, van Merri{\"e}nboer,
  Joulin, and Mikolov}]{weston2015towards}
Jason Weston, Antoine Bordes, Sumit Chopra, Alexander~M Rush, Bart van
  Merri{\"e}nboer, Armand Joulin, and Tomas Mikolov. 2015.
\newblock \href {https://arxiv.org/abs/1502.05698} {Towards {AI}-complete
  question answering: A set of prerequisite toy tasks}.
\newblock \emph{arXiv preprint arXiv:1502.05698}.

\bibitem[{Wolf et~al.(2020)Wolf, Debut, Sanh, Chaumond, Delangue, Moi, Cistac,
  Rault, Louf, Funtowicz, Davison, Shleifer, von Platen, Ma, Jernite, Plu, Xu,
  Le~Scao, Gugger, Drame, Lhoest, and Rush}]{wolf2019transformers}
Thomas Wolf, Lysandre Debut, Victor Sanh, Julien Chaumond, Clement Delangue,
  Anthony Moi, Pierric Cistac, Tim Rault, Remi Louf, Morgan Funtowicz, Joe
  Davison, Sam Shleifer, Patrick von Platen, Clara Ma, Yacine Jernite, Julien
  Plu, Canwen Xu, Teven Le~Scao, Sylvain Gugger, Mariama Drame, Quentin Lhoest,
  and Alexander Rush. 2020.
\newblock Transformers: State-of-the-art natural language processing.
\newblock In \emph{Proceedings of the 2020 Conference on Empirical Methods in
  Natural Language Processing: System Demonstrations}, pages 38--45.

\bibitem[{Xie et~al.(2020)Xie, Thiem, Martin, Wainwright, Marmorstein, and
  Jansen}]{xie2020worldtree}
Zhengnan Xie, Sebastian Thiem, Jaycie Martin, Elizabeth Wainwright, Steven
  Marmorstein, and Peter Jansen. 2020.
\newblock Worldtree v2: A corpus of science-domain structured explanations and
  inference patterns supporting multi-hop inference.
\newblock In \emph{Proceedings of The 12th Language Resources and Evaluation
  Conference}, pages 5456--5473.

\bibitem[{Xu et~al.(2018)Xu, Zhang, Qu, Xie, and Nock}]{xu2018d}
Qiongkai Xu, Juyan Zhang, Lizhen Qu, Lexing Xie, and Richard Nock. 2018.
\newblock D-page: Diverse paraphrase generation.
\newblock \emph{arXiv preprint arXiv:1808.04364}.

\bibitem[{Yang et~al.(2018)Yang, Qi, Zhang, Bengio, Cohen, Salakhutdinov, and
  Manning}]{yang2018hotpotqa}
Zhilin Yang, Peng Qi, Saizheng Zhang, Yoshua Bengio, William Cohen, Ruslan
  Salakhutdinov, and Christopher~D. Manning. 2018.
\newblock \href {https://doi.org/10.18653/v1/D18-1259} {{H}otpot{QA}: A dataset
  for diverse, explainable multi-hop question answering}.
\newblock In \emph{Proceedings of the 2018 Conference on Empirical Methods in
  Natural Language Processing}, pages 2369--2380, Brussels, Belgium.
  Association for Computational Linguistics.

\bibitem[{Ye et~al.(2020)Ye, Huang, Boschee, and Ren}]{ye2020teaching}
Qinyuan Ye, Xiao Huang, Elizabeth Boschee, and Xiang Ren. 2020.
\newblock Teaching machine comprehension with compositional explanations.
\newblock In \emph{Proceedings of the 2020 Conference on Empirical Methods in
  Natural Language Processing: Findings}, pages 1599--1615.

\bibitem[{Yu et~al.(2019)Yu, Chang, Zhang, and Jaakkola}]{yu2019rethinking}
Mo~Yu, Shiyu Chang, Yang Zhang, and Tommi Jaakkola. 2019.
\newblock Rethinking cooperative rationalization: Introspective extraction and
  complement control.
\newblock In \emph{Proceedings of the 2019 Conference on Empirical Methods in
  Natural Language Processing and the 9th International Joint Conference on
  Natural Language Processing (EMNLP-IJCNLP)}, pages 4085--4094.

\bibitem[{Zaheer et~al.(2017)Zaheer, Kottur, Ravanbakhsh, Poczos,
  Salakhutdinov, and Smola}]{Zaheer2017deepsets}
Manzil Zaheer, Satwik Kottur, Siamak Ravanbakhsh, Barnabas Poczos, Russ~R
  Salakhutdinov, and Alexander~J Smola. 2017.
\newblock \href
  {https://proceedings.neurips.cc/paper/2017/file/f22e4747da1aa27e363d86d40ff442fe-Paper.pdf}
  {Deep sets}.
\newblock In \emph{Advances in Neural Information Processing Systems},
  volume~30, pages 3391--3401. Curran Associates, Inc.

\bibitem[{Zaidan et~al.(2007)Zaidan, Eisner, and Piatko}]{zaidan2007using}
Omar Zaidan, Jason Eisner, and Christine Piatko. 2007.
\newblock Using “annotator rationales” to improve machine learning for text
  categorization.
\newblock In \emph{Human language technologies 2007: The conference of the
  North American chapter of the association for computational linguistics;
  proceedings of the main conference}, pages 260--267.

\bibitem[{Zettlemoyer and Collins(2005)}]{zettlemoyer2012learning}
Luke~S. Zettlemoyer and Michael Collins. 2005.
\newblock \href {https://dl.acm.org/doi/10.5555/3020336.3020416} {Learning to
  map sentences to logical form: Structured classification with probabilistic
  categorial grammars}.
\newblock In \emph{Proceedings of the Twenty-First Conference on Uncertainty in
  Artificial Intelligence}, UAI’05, page 658–666. AUAI Press.

\bibitem[{Zhang et~al.(2020)Zhang, Zhao, and Song}]{zhang2020WinoWhy}
Hongming Zhang, Xinran Zhao, and Yangqiu Song. 2020.
\newblock \href {https://doi.org/10.18653/v1/2020.acl-main.508} {{W}ino{W}hy: A
  deep diagnosis of essential commonsense knowledge for answering {W}inograd
  schema challenge}.
\newblock In \emph{Proceedings of the 58th Annual Meeting of the Association
  for Computational Linguistics}, pages 5736--5745, Online. Association for
  Computational Linguistics.

\bibitem[{Zhang et~al.(2019{\natexlab{a}})Zhang, Hare, and
  Prugel-Bennett}]{Zhang2019deepsets}
Yan Zhang, Jonathon Hare, and Adam Prugel-Bennett. 2019{\natexlab{a}}.
\newblock \href
  {https://proceedings.neurips.cc/paper/2019/file/6e79ed05baec2754e25b4eac73a332d2-Paper.pdf}
  {Deep set prediction networks}.
\newblock In \emph{Advances in Neural Information Processing Systems},
  volume~32, pages 3212--3222. Curran Associates, Inc.

\bibitem[{Zhang et~al.(2019{\natexlab{b}})Zhang, Hare, and
  Pr{\"u}gel-Bennett}]{zhang2019fspool}
Yan Zhang, Jonathon Hare, and Adam Pr{\"u}gel-Bennett. 2019{\natexlab{b}}.
\newblock Fspool: Learning set representations with featurewise sort pooling.
\newblock In \emph{International Conference on Learning Representations}.

\end{thebibliography}
\bibliographystyle{acl_natbib}

\appendix
\section{Appendix}
\begin{table*}[ht!]
\small
\centering
\begin{tabular}{lccccccccccc}
\toprule
           &    & \multicolumn{3}{c}{Node} & \multicolumn{3}{c}{Edge} & \multicolumn{4}{c}{Proof} \\ \cmidrule(lr){3-5} \cmidrule(lr){6-8} \cmidrule(lr){9-11}
           & QA & P      & R      & F1     & P      & R      & F1     & P    & R    & F1   & FA   \\ \midrule
\textsc{PRover} \cite{saha2020prover} & 99.3 & 90.0 & 85.3 & 86.4 & 88.6 & 85.6 & 86.2 & 88.0 & 84.7 & 85.5 & 82.1\\
\textsc{PRover}-all     & 99.4 &   88.4 &	84.2 &	85.3 &	87.9 &	84.4 &	85.4 &	86.5 &	83.5 &	84.3 &	81.1       \\
\textsc{PRover}-top-$p$ & 99.4 & 34.4 & 88.6 & 48.4 & 34.0 & 88.0 & 48.0 & 33.4 & 87.3 & 47.4 & 00.0     \\
\textsc{PRover}-top-$p$-classifier & 99.4 & 86.2 & 85.1 & 84.4 & 85.6 & 85.1 & 84.4 & 84.6 & 84.2 & 83.4 & 78.2   \\
\textsc{PRover}-top-$p$-threshold & 99.4 & 85.0 & 88.4 & 85.6 & 84.5 & 87.8 & 85.2 & 83.9 & 87.1 & 84.5 & 78.0  \\
\shortmodelmult{} & 99.3 &	89.9 &	89.6 &	89.3 &	88.3 &	88.3 &	88.0 &	87.7 &	87.6 &	87.3 &	83.9      \\
\shortmodelseq{}-seq & 99.4   & 88.5      & 86.8       &  87.2      & 87.4       &   86.0     &    86.3    &  86.7    &  85.3    &  85.6    &  81.4   \\
\shortmodelseq{}-nec & 99.3   & 90.2      & 89.7        &  89.5      & 89.4       &   89.1     &    89.0    &  89.0    &  88.6    &  88.5    &  85.2   \\
\shortmodelseq{} & \textbf{99.5} & \textbf{90.6} & \textbf{90.5} & \textbf{90.1} & \textbf{89.9} & \textbf{90.0} & \textbf{89.5} & \textbf{89.4} & \textbf{89.4} & \textbf{89.0} & \textbf{85.3} \\
\bottomrule
\end{tabular}
\caption{\label{du5-dev} Comparison of our final \model{} models with variants of \textsc{PRover} and other ablations of \shortmodelseq{} on the validation set of DU5. \shortmodelseq-seq = \modelseq{} with sequential loss. \shortmodelseq-nec = \modelseq{} with no edge conditioning. The final \modelseq{} model outperforms all other models across all metrics.}
\vspace{-5pt}
\end{table*}

\subsection{Experimental Setup}

\model{} is developed on top of the Hugging Face transformers library \cite{wolf2019transformers}.\footnote{\url{https://github.com/huggingface/transformers}} Experiments with \textsc{PRover} \cite{saha2020prover} are performed using their publicly released code and hyperparameters.\footnote{\url{https://github.com/swarnaHub/PRover}} All \model{} hyperparameters are chosen based on the best Full Accuracy on the corresponding validation set. We use RoBERTa-large \cite{liu2019roberta} as the pre-trained language model. The batch size and maximum sequence length are set to $8$ and $300$ respectively. We train all our models for a maximum of $7$ epochs using an initial learning rate of $10^{-5}$, a weight decay of $0.1$ and a dropout probability of $0.1$. We use a random seed of $42$ across all our experiments. All experiments are performed on one V100 Volta GPU. Batch size and learning rate are manually tuned in the range \{$8$,$16$\} and \{$10^{-5}$, $2*10^{-5}$\} respectively. For inference, we use \textsc{PRover}'s ILP optimization code, which is modeled using PuLP.\footnote{\url{https://pypi.org/project/PuLP/}} In all the datasets, the maximum number of facts and rules corresponding to a context is 25.

\subsection{Datasets}
Our experiments are conducted on the datasets introduced in \citet{clark2020transformers}.\footnote{\url{ https://rule-reasoning.apps.allenai.org/}} These consist of 5 datasets with synthetic rule-bases, DU0-DU5, a zero-shot test-only dataset called Birds-Electricity and a dataset with human-paraphrased rules called ParaRules. All datasets, except Birds-Electricity, have their corresponding train, validation and test splits.

\paragraph{\textbf{DU0-DU5: }}Each of these consists of 100k questions with synthetic rule-bases and requires reasoning chains up to a maximum depth of $D$ ($D=0,1,2,3,5$). The number of train, validation and test examples in each of the datasets are 70k, 10k and 20k respectively. Further, each question in the datasets is annotated with all possible proofs. The total number of proofs in the DU5 train set range from $1$ to $1350$, with a mean and median of $1.45$ and $1$ respectively.

\paragraph{\textbf{Birds-Electricity: }}The Birds-Electricity dataset comprises of two test-only datasets where the contexts are about birds and electric circuits. The vocabulary of the entities, attributes and predicates, apart from \texttt{is()} are all new at test time, thus providing a benchmark for testing the generalization capability of the models on out-of-distribution data. Another interesting aspect of this dataset is that all examples are annotated with a unique gold proof.

\paragraph{\textbf{ParaRules: }}The ParaRules dataset is one where the facts and rules are paraphrased by humans into more natural language. It consists of a total of 40k questions, with 28k, 4k, and 8k questions in the train, validation and test splits respectively. This dataset tests the model's ability to reason over more complex human-like language. Similar to the synthetic datasets, each example is annotated with all possible proofs.

\begin{table*}[t]
\small
\centering
\begin{tabular}{lccccccccccc}
\toprule
           &    & \multicolumn{3}{c}{Node} & \multicolumn{3}{c}{Edge} & \multicolumn{4}{c}{Proof} \\ \cmidrule(lr){3-5} \cmidrule(lr){6-8} \cmidrule(lr){9-12}
        p   & QA & P      & R      & F1     & P      & R      & F1     & P    & R    & F1   & PA   \\ \midrule
2 & 99.4 & 90.5 & 89.1 & 89.2 & 89.3 & 88.3 & 88.4 & 88.8 & 87.8 & 87.9 & 84.4    \\
3 & 99.3 &	89.9 &	89.6 &	89.3 &	88.3 &	88.3 &	88.0 &	87.7 &	87.6 &	87.3 &	83.9      \\
4 & 99.3 & 89.1 & 89.2 & 88.8 & 87.8 & 82.0 & 87.8 & 87.2 & 87.5 & 87.1 & 83.6 \\ 
5 & 99.2 & 88.6 & 89.1 & 88.5 & 87.2 & 87.8 & 87.2 & 86.6 & 87.2 & 86.6 & 83.1 \\ \bottomrule  
\end{tabular}
\caption{\label{label-exp-ablation} Effect of varying maximum number of proofs (p) on \modelmult{}. All models are trained on the DU5 training set and evaluated on the corresponding validation set. The proof metrics start to decrease marginally with increase in p.}
\end{table*}

\begin{table*}[t]
\small
\centering
\begin{tabular}{lccccccccccc}
\toprule
           &    & \multicolumn{3}{c}{Node} & \multicolumn{3}{c}{Edge} & \multicolumn{4}{c}{Proof} \\ \cmidrule(lr){3-5} \cmidrule(lr){6-8} \cmidrule(lr){9-12}
        p   & QA & P      & R      & F1     & P      & R      & F1     & P    & R    & F1   & PA   \\ \midrule
2 & 99.5 & 90.0 & 89.0 & 89.0 & 89.2 & 88.4 & 88.3 & 88.6 & 87.8 & 87.7 & 84.1   \\
3 & 99.5 & 90.6 & 90.5 & 90.1 & 89.9 & 90.0 & 89.5 & 89.4 & 89.4 & 89.0 & 85.3	    \\
4 & 99.5 & 90.2 & 89.7 & 89.5 & 89.5 & 89.2 & 89.1 & 89.1 & 88.7 & 88.6 & 85.2 \\
5 & 99.5 & 90.1 & 89.6 & 89.4 & 89.5 & 89.2 & 89.1 & 89.0 & 88.6 & 88.5 & 85.2  \\ \bottomrule  
\end{tabular}
\caption{\label{model-exp-ablation} Effect of varying maximum number of proofs (p) on \modelseq{}. All models are trained on the DU5 training set and evaluated on the corresponding validation set. Unlike \modelmult{}, it is significantly robust to variation in p.}
\end{table*}

\subsection{Syntax of Proof Graph}

Each proof $\mc{P}_i = (\mc{V}_i, \mc{E}_i)$ is a directed graph, with a set of nodes $\mc{V}_i \subseteq \mc{N}$ and a set of edges $\mc{E}_i \subseteq \mc{V}_i \times \mc{V}_i$. Each node $n_i \in \mc{N}$ is either a fact $F \in \mc{F}$ or a rule $R \in \mc{R}$ from the context or a special \textbf{NAF} node, denoting ``Negation as Failure". A \textbf{NAF} node in a proof indicates the truthfulness of the negation of statement(s) that
cannot be proved using the set of rules (under closed-world assumption). Edges in the graph can be directed either from a fact (or \textbf{NAF}) to a rule or between two rules. An edge from a fact to a rule means that the rule applies on the fact to generate a new fact. Similarly, an edge from a rule $R_1 \in \mc{R}$ to another rule $R_2 \in \mc{R}$ implies the application of $R_2$ on the fact generated by $R_1$. Proofs are either successful or failed. A successful proof is one where the question statement can be logically reached (to be either proved or disproved) using the given rule-base while for failed proofs, no conclusion can be reached, in which case the shallowest branch of the proof tree that fails is generated. For more details and examples of proofs, we refer the readers to prior work \cite{saha2020prover, clark2020transformers}.

\subsection{Ablation Analysis}

In Table \ref{du5-dev}, we compare our baselines \textsc{PRover}, \textsc{PRover}-all and \textsc{PRover}-top-$p$ variants with our \model{} models on the validation set of DU5 dataset. Additionally, we also show two ablations of \shortmodelseq{} -- in the first, we replace the Hungarian loss with a sequential loss, which computes the cross-entropy loss of the $i^{th}$ predicted proof with the $i^{th}$ gold proof and in the second, we condition the node embeddings on the previous node embeddings only instead of both node and edge embeddings. Except \textsc{PRover} and \textsc{PRover}-all, which generate a single proof, all other models generate a maximum of 3 proofs. \textsc{PRover}-top-$p$ suffers from a significant drop in proof precision due to the generation of many incorrect proofs. Although carefully choosing the value of $p$ either by thresholding or through a classifier helps boost the proof precision, \textsc{PRover} continues to be a superior baseline on this dataset due to a high skew towards single-proof examples. \shortmodelmult{} improves upon \textsc{PRover}'s proof F1 and full accuracy (FA) which are further bettered by \shortmodelseq{}, owing to its explicit conditioning mechanism between the proofs. Replacing the Hungarian loss with a sequential loss leads to a significant drop in proof F1, thereby showing the effectiveness of modeling multiple proof generation as a set generation problem. Finally, conditioning the node embeddings on both node and edge embeddings leads to marginal improvement in proof F1. Overall, \shortmodelseq{} outperforms all other models across all metrics.

\begin{table*}[t]
\small
\centering
\begin{tabular}{lccccccccccc}
\toprule
           &    & \multicolumn{3}{c}{Node} & \multicolumn{3}{c}{Edge} & \multicolumn{4}{c}{Proof} \\ \cmidrule(lr){3-5} \cmidrule(lr){6-8} \cmidrule(lr){9-11}
           & QA & P      & R      & F1     & P      & R      & F1     & P    & R    & F1   & FA   \\ \midrule
\textsc{PRover}-all     &  98.6 & 95.9 & 94.1 & 94.5 & 95.4 & 93.8 & 94.3 & 95.3 & 93.7 & 94.2 & 92.3   \\
\textsc{PRover}-top-$p$ & 98.6 & 39.3 & 96.6 & 55.0 & 38.9 & 96.0 & 54.6 & 38.9 & 95.9 & 54.5 & 00.1 \\
\shortmodelmult{} & \textbf{98.9} & 96.7 & 96.4 & 96.4 & 96.4 & 96.2 & 96.2 & 96.2 & 96.0 & 96.0 & 95.2 \\
\shortmodelseq{}  & \textbf{98.9} & \textbf{97.3} & \textbf{97.2} & \textbf{97.2} & \textbf{97.2} & \textbf{97.0} & \textbf{97.0} & \textbf{96.8} & \textbf{96.7} & \textbf{96.7} & \textbf{96.1} \\
\bottomrule
\end{tabular}
\caption{\label{natlang-dev} Comparison  of  models  trained  on  DU3  and ParaRules training sets and evaluated on ParaRules validation set. \shortmodelseq{ outperforms all other models across all metrics.}}
\end{table*}

\begin{table*}[t]
\small
\centering
\begin{tabular}{lccccccccccc}
\toprule
           &    & \multicolumn{3}{c}{Node} & \multicolumn{3}{c}{Edge} & \multicolumn{4}{c}{Proof} \\ \cmidrule(lr){3-5} \cmidrule(lr){6-8} \cmidrule(lr){9-11}
           & QA & P      & R      & F1     & P      & R      & F1     & P    & R    & F1   & FA   \\ \midrule
\textsc{PRover}-all     &  98.2 & 95.3 & 92.8 & 93.5 & 94.7 & 92.7 & 93.3 & 94.4 & 92.4 & 93.0 & 90.5   \\
\textsc{PRover}-top-$p$ & 98.2 & 38.7 & 95.9 & 54.3 & 38.3 & 95.5 & 53.9 & 38.2 & 95.3 & 53.8 &  00.1  \\
\shortmodelmult{} & \textbf{98.3} & 96.0 & 95.6 & 95.7 & 95.9 & 95.5 & 95.6 & 95.5 & 95.2 & 95.2 & 93.8  \\
\shortmodelseq{} & \textbf{98.3} & \textbf{96.8} & \textbf{96.2} & \textbf{96.3} &  \textbf{96.5} & \textbf{96.3} & \textbf{96.3} & \textbf{96.2} & \textbf{96.0} & \textbf{96.0} & \textbf{94.5} \\
\bottomrule
\end{tabular}
\caption{\label{natlang-test} Comparison  of  models  trained  on  DU3  and ParaRules training sets and evaluated on ParaRules test set. \shortmodelseq{ outperforms all other models across all metrics.}}
\end{table*}

\subsection{\model{} with Varying Maximum Number of Proofs}

We analyze the effect of varying the maximum number of proofs $p$ on \shortmodelmult{} and \shortmodelseq{} in Table \ref{label-exp-ablation} and \ref{model-exp-ablation} respectively. All models are trained on the DU5 training set and evaluated on the corresponding validation set. Although all models obtain high QA accuracy, we observe that the proof F1 for \shortmodelmult{} starts to decrease marginally with the increase in $p$. Note that this model is trained with padding of empty proof graphs since it generates all $p$ proofs in parallel. Thus, the amount of padding increases with the increase in $p$, which in turn requires the model to predict more empty proof graphs, thereby leading to a harder learning problem. \shortmodelseq{}, on the other hand, is significantly robust to such variations in $p$, because it generates proofs iteratively with one empty graph at the end, indicating end of generation.

\subsection{Evaluation on Human-Paraphrased Rule-Bases}
Following \textsc{PRover}, we also test \model{}'s effectiveness in generating proofs for more human-like complex rule-bases. The ParaRules dataset is constructed by first creating a set of fact groups where each fact group consists of all facts in the theory concerning a particular person and then paraphrasing these fact groups into more complex language. E.g., a fact group ``Alan is blue.  Alan is rough.  Alan is young.", can be re-worded into ``Alan is on the young side, but rough. He often feels rather blue." Thus, unlike the DU datasets or the Birds-Electricity dataset where the proof graphs are composed of facts and rules, ParaRules proofs are composed of fact groups and rules. Following past work \cite{clark2020transformers, saha2020prover}, we train our models combining the DU3 and ParaRules train sets, and evaluate on the ParaRules validation and test set in Tables \ref{natlang-dev} and \ref{natlang-test} respectively. We find that similar conclusions to the DU5 dataset hold for this dataset as well -- \shortmodelmult{} achieves a better proof F1 and full accuracy than \textsc{PRover}, which are further improved by \shortmodelseq{} due to its explicit conditioning mechanism between the proofs.

\begin{table}
\small
\centering
\begin{tabular}{lccccccc}
\toprule
           & \multicolumn{3}{c}{\# Parameters} & \multicolumn{3}{c}{Time/epoch (in hours)} & \\ \cmidrule(lr){2-4} \cmidrule(lr){5-7}
           p & PR & ML      & IT      & PR     & ML      & IT   \\ \midrule
           1 & 361M & 361M & 488M & 5.0 & 3.4 & 3.6\\
           2 & 361M & 361M & 615M & 5.0 & 3.5 & 4.0\\
           3 & 361M & 361M & 742M & 5.0 & 3.6 & 4.6\\
           4 & 361M & 361M & 869M & 5.0 & 3.7 & 5.1\\
           5 & 361M & 361M & 996M & 5.0 & 3.8 & 5.7\\
\bottomrule
\end{tabular}
\caption{\label{size-time} Comparative study of the number of parameters and training time per epoch (in hours) for \textsc{PRover}-all (PR), \shortmodelmult{} and \shortmodelseq{} with varying number of maximum proofs (p).}
\end{table}

\subsection{Training Time and Size Comparison}
Table \ref{size-time} shows the number of trainable parameters and training times per epoch for the baseline model \textsc{PRover} and our proposed models, \shortmodelmult{} and \shortmodelseq{} across varying number of maximum proofs ($p$) per sample. Since \shortmodelmult{} adopts the same \textsc{PRover} architecture but with multi-label classification, it has the same number of parameters as \textsc{PRover}, which also remains unchanged irrespective of the maximum number of proofs. The number of parameters for  \shortmodelseq{}, however, increases with the increase in $p$ because of the presence of multiple node and edge encoders. While \shortmodelseq{} has more parameters than \textsc{PRover}, our empirical findings reveal that just having a similarly-sized, larger \textsc{PRover} model will not be sufficient and exploiting the correlations between multiple proofs with a permutation-invariant loss is necessary for the task of generating a set of multiple proofs.

The training time of \textsc{PRover} is more than that of \shortmodelmult{} because the former treats each proof as a separate example, causing an increase in the training data size from 70k to 110k. \shortmodelmult{} is the most time-efficient model and its running time only increases marginally with the increase in $p$. This is due to the additional node and edge classifications that the model has to perform corresponding to each additional proof. Unsurprisingly, \shortmodelseq{} takes longer to train but encouragingly for $p \leq 4$, still has a comparable running time to \textsc{PRover}.

\end{document}